\documentclass[preprint,12pt]{elsarticle}

\usepackage{amssymb}
\usepackage{amsmath}
\usepackage{booktabs}
\usepackage{subcaption}
\usepackage{algorithm}
\usepackage{algpseudocode}

\journal{Expert Systems with Applications}

\begin{document}

\begin{frontmatter}



\title{Deep Learning-Based Control Optimization for Glass Bottle Forming}


\author[fbk]{Mattia Pujatti}
\ead{mpujatti@fbk.eu}
\author[fbk]{Andrea Di Luca} 
\ead{adiluca@fbk.eu}
\author[glassform]{Nicola Peghini}
\ead{nicola.peghini@glassform.ai}
\author[glassform]{Federico Monegaglia}
\ead{federico.monegaglia@glassform.ai}
\author[fbk]{Marco Cristoforetti} 
\ead{marco.cristoforetti@fbk.eu}

\affiliation[fbk]{%
    organization={Fondazione Bruno Kessler},%
    addressline={Via Sommarive 18},%
    city={Trento},%
    postcode={38123},%
    state={TN},%
    country={Italy}%
}

\affiliation[glassform]{%
    organization={GLASSFORM.AI S.p.A.},%
    addressline={Piazza Manifattura 1},%
    city={Rovereto},%
    postcode={38068},%
    state={TN},%
    country={Italy}%
}

\begin{abstract}
In glass bottle manufacturing, precise control of forming machines is critical for ensuring quality and minimizing defects. This study presents a deep learning-based control algorithm designed to optimize the forming process in real production environments. Using real operational data from active manufacturing plants, our neural network predicts the effects of parameter changes based on the current production setup. Through a specifically designed inversion mechanism, the algorithm identifies the optimal machine settings required to achieve the desired glass gob characteristics. Experimental results on historical datasets from multiple production lines show that the proposed method yields promising outcomes, suggesting potential for enhanced process stability, reduced waste, and improved product consistency.  These results highlight the potential of deep learning to process control in glass manufacturing.
\end{abstract}




\end{frontmatter}


\section{Introduction}
\label{sec:intro}

Despite growing interest in Artificial Intelligence (AI) applications, their integration into industrial manufacturing processes is still limited. This is due to a combination of factors. Technologically, many systems lack the necessary infrastructure for effective sensing, data acquisition, and real-time processing. At the same time, the inherent complexity of describing these industrial processes poses significant challenges for the successful implementation of modern data-driven algorithms.

This is particularly evident in glass manufacturing, where fluid dynamics are governed by the Navier-Stokes equations, which are notoriously challenging to model accurately, especially for atypical glass compositions \cite{https://doi.org/10.1111/ijag.15881}. As a result, manufacturers tend to rely on dedicated simulators, which are often computationally and economically expensive and complex to adapt to different environments \cite{MOURTZIS2014213, simulation_glass_technology}. 
The tuning of machine settings is usually entrusted to technicians, bounding this with the specific expertise and sensitivity of the operator.

In this work, we propose an AI-driven machine control framework designed to augment human expertise in managing the feeder mechanism of glass production lines. Our solution comprises two components: a deep learning model trained to replicate the behavior of the forming machine, and an inversion algorithm designed to identify the optimal settings that meet specific production targets. 
The inversion mechanism we employ—training a forward neural-network model of the process and subsequently optimizing its inputs through Monte Carlo or gradient-based methods—is a well-established approach in other industrial domains such as furnace control and materials inverse design \cite{Yoon2025Materials, Yadav01092020}.
However, our contribution lies in adapting and validating this established inversion paradigm for the specific context of glass bottle manufacturing, and demonstrating that it can operate effectively on historical production data from real plants.
This represents a substantial challenge, as the fluid dynamics of molten glass are highly complex and strongly dependent on local boundary conditions, making accurate physical simulation prohibitively expensive and requiring fine-tuned, process-specific simulators.

In contrast to most existing inversion applications, where forward models are trained on data generated from explicit physical equations or numerical simulators, our framework is entirely data-driven.
It is trained directly on historical production data, enabling it to capture the real variability of industrial operations rather than relying on synthetic or simulated samples.

This approach enables the automatic adjustment of machine parameters to achieve the desired glass gob characteristics, thereby improving production accuracy, minimizing material waste, reducing downtime, and enhancing both operational stability and workplace safety.
The deep learning component contributes to model generalizability across different product types and conditions; the inversion algorithm, on the other hand, provides valuable insights into the complex relationships between machine parameters and product properties, offering both predictive and prescriptive capabilities within a unified framework.

This paper is structured as follows. Section~\ref{sec:related_works} reviews current strategies employed in automatic machine control within the glass manufacturing industry, highlighting their limitations compared to the proposed approach. Then, Section~\ref{sec:experimental_procedure} provides a detailed description of  the framework employed for the analysis and development of this solution, accompanied by the characteristics of the datasets used. This section also introduces the proposed methodology, including a thorough explanation of both the deep learning model and the inversion procedure. Subsequently, Section~\ref{sec:results_discussion} presents and discusses the results obtained by applying the algorithm to a real-world case study. Finally, Section~\ref{sec:conclusions} summarizes the key contributions of this work and outlines directions for future research.

\section{Related works}
\label{sec:related_works}
Despite being a core part of hollow glass production, the control of forming machines, particularly the Independent Section (IS) machine, has remained largely manual or reliant on proprietary, undocumented industrial solutions. Many facilities still involve human operators manually adjusting machine settings to comply with production requests. These parameters are usually critical in shaping the final container geometry but cannot be tuned in real-time due to the nature of the process: many defects manifest hours later at the cold end of the line, making direct cause-and-effect relationships difficult to model and correct. This delay between intervention and feedback introduces significant uncertainty and limits opportunities for online optimization. As a result, production efficiency and quality remain highly dependent on operator expertise and experience.

Where automation is implemented, it typically relies on proportional-integral-derivative (PID) controllers \cite{AstromHagg95}, a feedback-based mechanism widely used in general industrial manufacturing, whose accuracy is inevitably affected by the aforementioned delays. More sophisticated solutions include model predictive control (MPC) \cite{mpc_glass_furnaces}, which optimizes processes under constraints by incorporating future model predictions. Intermediate approaches also exist, such as \cite{cv_weight_gob_control}, which are model-informed yet do not exploit future predictions. However, these methods turn out to suffer from complex tuning requirements, particularly under highly dynamic operating conditions, such as those typical of glass manufacturing, illustrated in the case study analyzed.

Many of the more advanced automation strategies are unfortunately implemented as closed-source proprietary solutions \cite{KOVACEC2010304} with limited technical documentation or transparency. As such, they remain inaccessible for replication, benchmarking or academic evaluation.

Parallel research has explored AI applications in related areas of glass manufacturing. For instance, several works have developed AI-based vision systems for defect monitoring and detection in downstream quality control \cite{app122111192, 10109368}. Another line of research envisions digital twins of entire production systems \cite{8082476} to support real-time decision making. However, such approaches are often unnecessarily complex and practically unfeasible without a robust sensor infrastructure. To the best of our knowledge, no existing work applies deep learning directly to the control of IS machines during the glass forming stage.

One of the core barriers in the development and validation of such control systems is the lack of publicly available, high-fidelity simulators capable of reflecting the nonlinear, thermo-mechanical dynamics of glass forming. Existing industrial solutions may be complete, but are often expensive, closed patents or proprietary software \cite{NOGRID_pointsBlow, FORGE_Glass}, and therefore not suitable for broad adoption in any setup. This hampers the development of data-driven approaches that require either large volumes of historical production data or simulated environments.

This work introduces a deep learning-based control algorithm for IS machines, addressing these gaps. Our approach is lightweight, allowing online inference and control within production timescales. While focused on hollow glass forming, the architecture is designed to generalize to other multi-input, multi-output (MIMO) process control tasks with similar dynamics. In contrast to industrial black-box solutions, our method is fully described and intended as a research foundation for further development. We believe this is the first published attempt to bring deep learning to the heart of IS machine control, and to make such technology transparent and accessible to both industry and academic communities.

\section{Experimental Procedure}
\label{sec:experimental_procedure}
In this study, the primary experimental objective is to apply the proposed algorithm to the feeder of IS machines for managing and regulating the formation of glass gobs, to assess its feasibility in practice.
This system is intended to be integrated into the upper section of the forming machine, specifically the feeder, to facilitate computerized adjustment of the key parameters that influence the physical properties of the produced gobs.

\subsection{Glass Gob Forming Overview}
\label{subsec:experimental_setup}

\begin{figure}
    \centering
    \begin{minipage}{0.45\textwidth}
        \centering
        \includegraphics[width=0.9\textwidth]{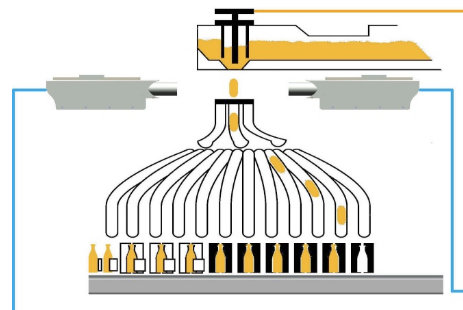}
    \end{minipage}
    \begin{minipage}{0.45\textwidth}
        \centering
        \includegraphics[width=0.9\textwidth]{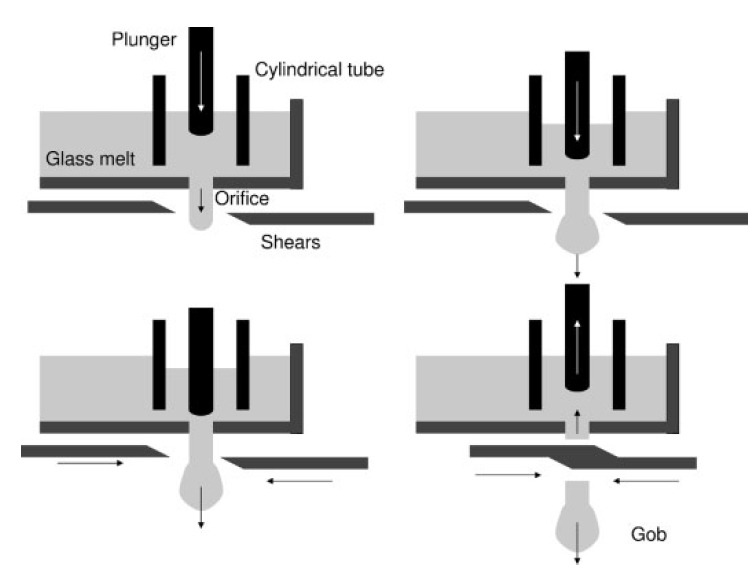}
    \end{minipage}\hfill
    \caption{Schema of the hollow glass forming machine (left) and focus on the four-step gob feeder process (right) \cite{glass_book}.}
    \label{fig:ismachine_feeder_schemas}
\end{figure}

The industrial core of hollow glass production addressed in this study consists primarily of two components: the feeder system and the Independent Section (IS) machine. The feeder, positioned at the top of the forming unit, is a servo-assisted mechanism responsible for shaping molten glass into uniform gobs with specific physical and geometric characteristics in terms of weight, length, and shape. The IS machine is located directly beneath the feeder and it is typically composed of several sections between 6 and 12. It is designed to receive those gobs and direct them into the corresponding blank molds (Figure~\ref{fig:ismachine_feeder_schemas}). These molds initiate the shaping phase of the forming process by giving the molten glass its first form. As the name suggests, each section of the IS machine operates independently, allowing for the simultaneous production of multiple containers.

The process begins with the molten glass entering the feeder through a rotating tube, which regulates its flow and keeps it homogeneous. Subsequently, a plunging mechanism cyclically extracts and propels controlled amounts of glass through an orifice, where a shear system precisely segments the flow into uniform droplets.

\begin{figure}
    \centering
    \includegraphics[width=\textwidth]{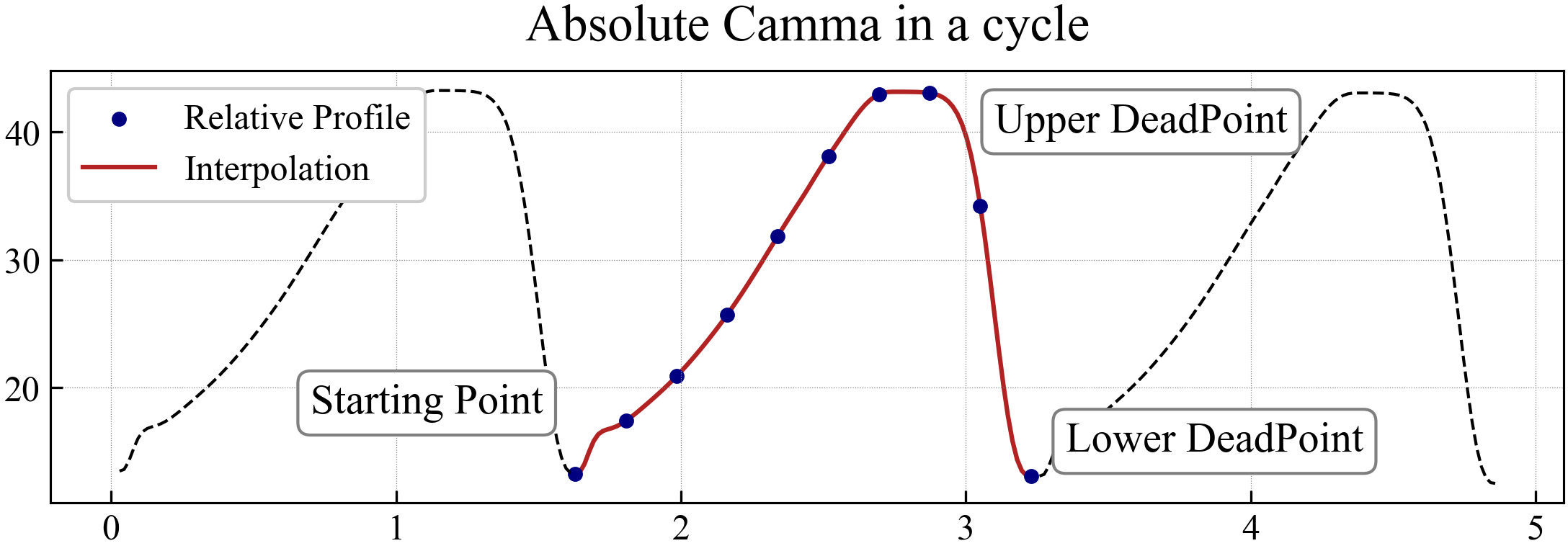}
    \caption{Example of a plunger movement profile within a cycle. Each Cam is constrained by its neighbors to comply with the mechanical constraints of the apparatus.}
    \label{fig:cam_cycle_example}
\end{figure}

In particular, the motion of the plungers plays a key role in regulating the glass flow and shaping the gobs according to specific requirements. These plungers operate in a vertical cyclic pattern, alternating between suction and pumping phases to control the passage of molten glass through the orifices (Figure~\ref{fig:ismachine_feeder_schemas}). The relative impact of each phase can vary significantly depending on the viscosity of the glass, and therefore, accurate control over the motion profile, known as \textit{Cam}, becomes crucial for the correct functioning of the machine. The actual motion of the plunger occurs solely along the vertical axis, therefore, the Cam profile, an example of which is given in Figure~\ref{fig:cam_cycle_example}, should be interpreted as a velocity diagram. In this sense, the y-axis corresponds to the actual vertical position of the plunger, and the x-axis is the elapsed time between consecutive up-down motions over different sections.
Precise control over gobs' characteristics is essential, as their dimensions must be compatible with the molds they are delivered to avoid defects in the final product. Particularly relevant are their shapes and weights.

Traditionally, configuring the system settings to accommodate production requirements has relied on time-consuming, tedious, and potentially hazardous manual procedures. Operators were required to adjust each component individually, despite the strong interdependence among these mechanisms, often necessitating numerous trial-and-error iterations to reach satisfactory results. Some automation attempts have employed conventional methods such as PID and physics-informed MPC (both introduced in Section~\ref{sec:related_works}). Although these approaches provide high precision in parameter selection, they adapt poorly to the highly dynamic nature of the system, particularly to variations in temperature, glass composition, or inevitable component consumption. As a result, they require frequent recalibration, leading to significant waste in time and material.

Regarding the data acquisition phase, the measurement system integrated within the industrial setup consists of a set of infrared cameras positioned directly beneath the mechanical shears, between the feeder and the IS machine. These cameras enable continuous monitoring of the glass gobs during their free fall toward the distribution system by acquiring their image immediately after the cut. The system can estimate key gob properties such as weight, volume, and length by applying a contour extrapolation algorithm to the captured images and integrating the resulting data with temperature measurements. 

Some production facilities integrate even more advanced measurement systems that utilize pairs of cameras arranged in stereo configurations, enabling three-dimensional reconstruction of the falling gobs. This approach allows for more accurate estimation of their geometric and physical characteristics.

\subsection{Data Analysis and processing}
\label{subsec:data_analyis_and_processing}

Data used for model development were provided by a glass bottle manufacturing company \cite{glassform} and comprise the sensor records spanning the period from May 2023 to February 2025. The dataset includes samples from two different production facilities, A and B, consisting of a total of four distinct production lines. For clarity, these lines are labeled as A-1, A-2, and A-3 for the first facility, and B-1 for the single line in the second facility.

The data consists of measurements acquired from the infrared sensors previously described in Section~\ref{subsec:experimental_setup}, temporally aligned with the machine settings that generated gobs with such characteristics. This alignment should enable the development of a model capable of learning the underlying relationships between machine parameters and the resulting physical properties of the gobs.

The forming machine operates in repeating cycles to take advantage of all available sections, thereby enabling the simultaneous production of multiple items and optimizing industrial throughput. During each cycle, the plunger's motion must comply with the mechanical constraints of the machine. 

The Cam profile used during production is chosen from a set of precomputed motion profiles (relative profile) provided by the machine manufacturer, and remains fixed throughout the production run. Different bottle shapes can be produced in each section by adjusting the deadpoints of the profile, which shifts the extreme positions of the plunger motion. Other than the relative profile, each Cam is defined by three key points: the Starting Point, corresponding to the vertical position (Y-coordinate) of the plunger at the beginning of the motion; the lower deadpoint, which represents the end of the downward stroke; and the upper deadpoint, that is the maximum height the plunger reaches during the motion. To ensure continuous and synchronized operation across sections, the starting point of a section must match the lower deadpoint of the preceding section. To ensure the feasibility of its dynamics, the feeder system applies an interpolation to the nominal Cam profile. This approach allows for the smooth and continuous connection of the motion curves across the sections, ensuring uninterrupted operations and maintaining a seamless gob forming process, as shown in Figure~\ref{fig:cam_cycle_example}.

\begin{figure}
    \centering
    \includegraphics[width=\linewidth]{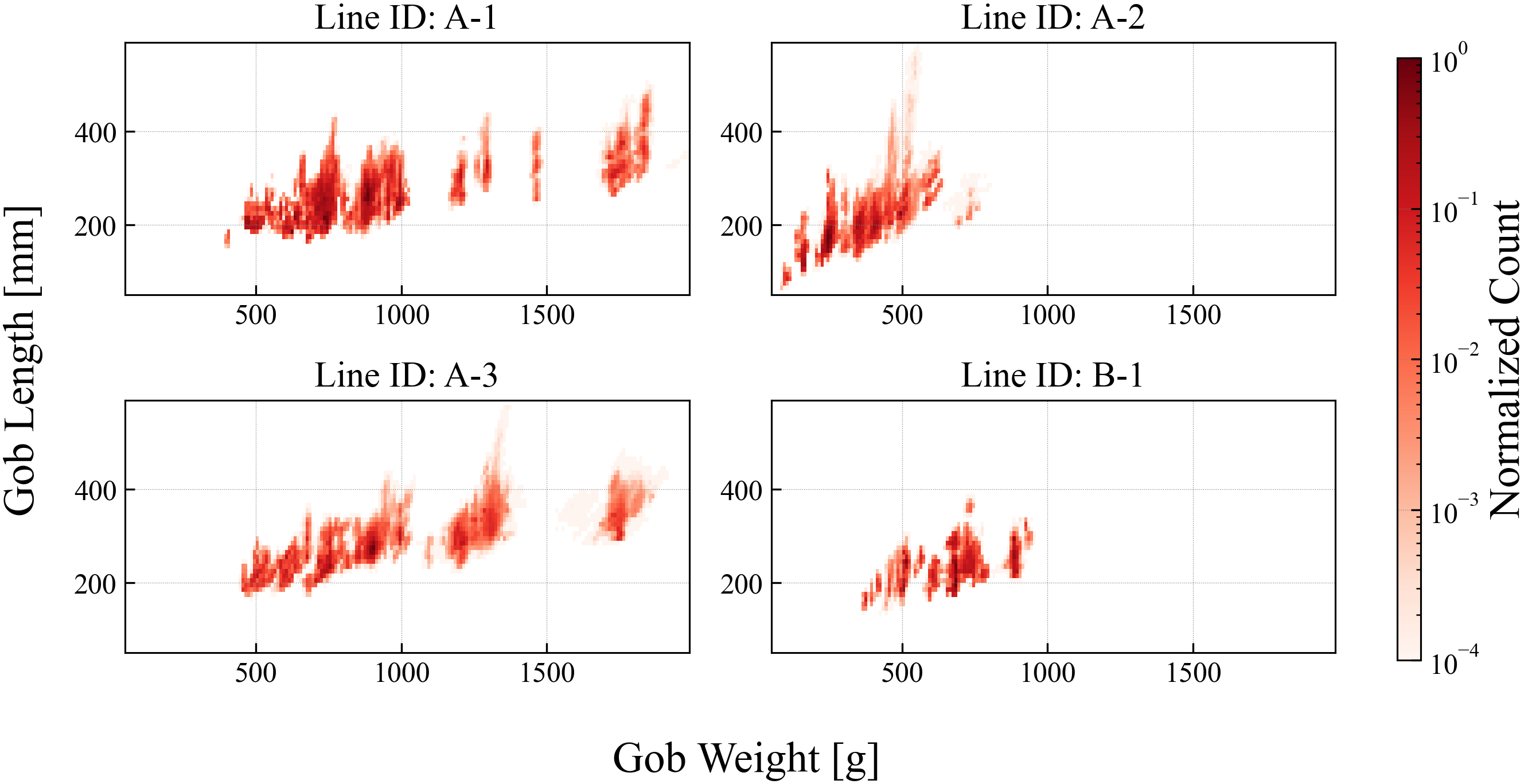}
    \caption{Distributions of the Weight-Length working points of the available lines.}
    \label{fig:wl_distributions}
\end{figure}

\begin{figure}
    \centering
    \includegraphics[width=\linewidth]{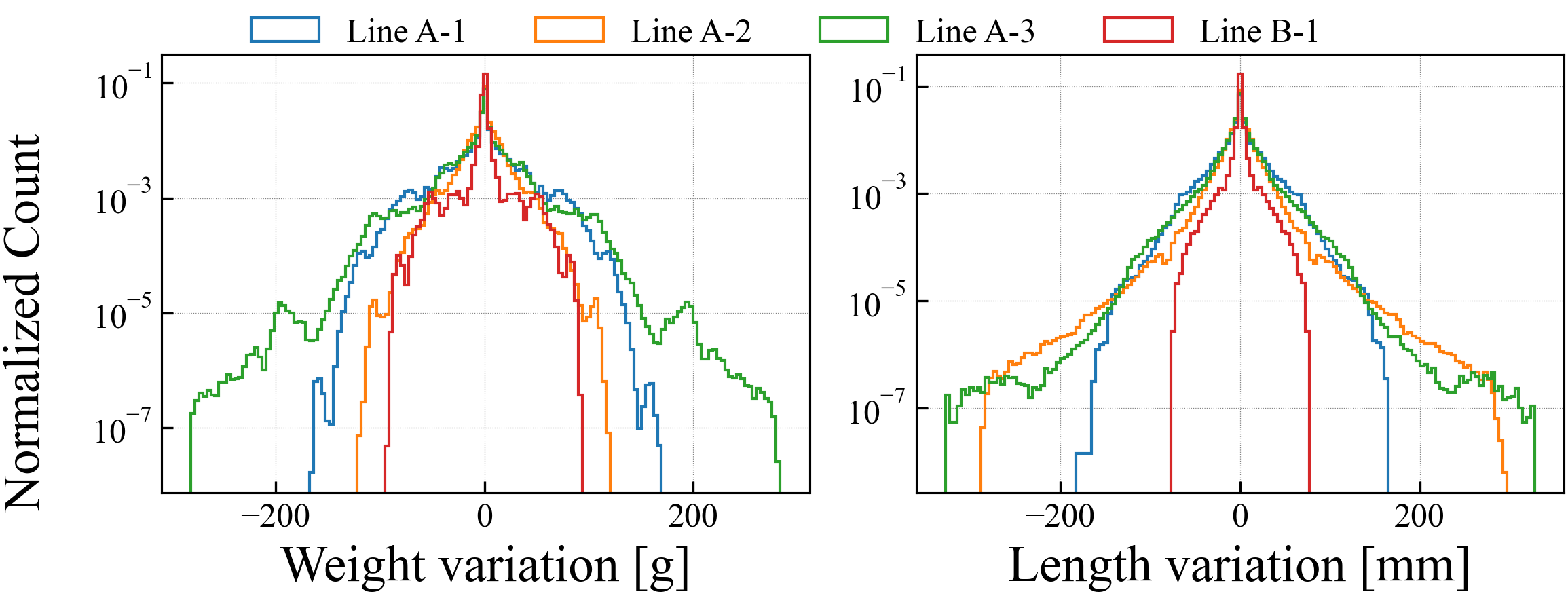}
    \caption{Distributions of the differences, in term of Weight and Length, between the sections within a same production. Hereafter, such variations will be used as targets by the regression model to learn general gobs transformations.}
    \label{fig:dwdl_distributions}
\end{figure}

The primary features targeted by the algorithmic control setup are the weight and length of the gobs produced in each section, which must be precisely adjusted to ensure compatibility with the corresponding molds and to meet customer specifications. Figure~\ref{fig:wl_distributions} illustrates the weight-length distribution of the available measurements, categorized by production line. Each production line generally operates around a specific working point, consistently producing bottles with similar characteristics. Figure~\ref{fig:dwdl_distributions} shows the variability between the features of gobs produced simultaneously in the same cycle.

When all gobs produced during a cycle are destined to the same mold type, the machine operates in "single-weight" mode. In this configuration, process parameters tend to be uniform across sections. Conversely, in "multi-weight" cycles, where gob weights vary between sections, it becomes essential to differentiate between global parameters, such as the feeder tube height, which influences all sections, and section-specific parameters, such as Cam boundary points, which must be individually adjusted.

Among the various machine parameters, the height of the feeder tube is undoubtedly the most critical to configure. This parameter directly influences the flow rate of molten glass through the orifice and is thus proportional to the average weight of the gobs produced. Once this global setting is established, fine-tuning the specific weight and length of gobs in each section can be achieved by adjusting the deadpoints.

In addition to these, a preliminary analysis has identified several other parameters that significantly affect the final characteristics of the gobs. These include the temperature of the molten glass, which affects its viscosity; the rotation speed of the feeder tube; the master speed, defined as the number of cuts per section made by the machine every minute; and the phase, which controls the timing offset between the plunger’s motion and the cutting operation.

The inclusion of different production lines and plans in the dataset is particularly important for demonstrating the model's ability to generalize effectively. As shown earlier, these lines tend to specialize in distinct products, which may vary considerably in shape, weight and length. For instance, as illustrated in Figure~\ref{fig:wl_distributions}, line A-2 focuses primarily on smaller bottles, whereas lines A-1 and A-3 operate across a much broader range. This distinction is generally reflected as well in the chemical composition of the glass processed. From a mechanical standpoint, production lines and plants also differ in the number of sections and cavities, and in the plunger type, resulting in varying production efficiencies and more or less compact Cam profiles.

\subsubsection{Differential approach}
\label{subsubsec:differential_approach}

The core idea of the proposed algorithm is to begin from the current production configuration on the facility and exploit the previously described setup to regulate the feeder tube height, thereby fixing the working point of the line. Once this global adjustment is completed, a deep learning model is employed to fine-tune the Cam deadpoints to precisely achieve the desired characteristics of the gobs over each section. 

To accomplish this, it is essential to develop a model capable of learning the actual mapping between adjustments to the cam profiles and the resulting variations in gob weight and length, as measured by the sensors on each section in response to those changes.

The dataset is constructed by collecting, for each production cycle, the differences between a reference section and another one randomly selected. This process yields two sets of variables: global settings, which describe the overall machine state, and feature-differential variables, representing the individual adjustments of each section. Due to the nature of this differential approach, the resulting data points tend to be highly similar, especially for single-weight cycles. To mitigate this issue, data resampling and duplicate removal strategies are applied during the training phase. 

The underlying assumption is that the selected global features, listed in Section~\ref{subsec:data_analyis_and_processing}, are sufficient to characterize the machine's operational state uniquely, such that, once fixed, the model can effectively learn to map the adjustments in the Cam deadpoints to the corresponding local variations in gobs' weights and lengths.

The data preprocessing step involves a systematic cleaning of missing or non-physical values, such as negative velocities or unrealistic weights. In addition to this standard filtering, a more dynamic approach was developed to identify stable clusters of data points and remove outliers within them, further enhancing the consistency and reliability of the samples used for model training.

\subsection{Methods}
\label{subsec:methods}

As explained in Section~\ref{subsubsec:differential_approach}, the proposed algorithm is designed to provide the necessary adjustments to the Cam deadpoints to achieve specific variations in gob weight and length for each section, given a set of operating conditions such as glass temperature and production speed.

A key challenge in this task lies in the underlying non-injective nature of the direct relationship between the variables involved in the study:
\begin{equation}
    \texttt{machine state} + \Delta W, \Delta L \to \Delta\texttt{deadpoints}
    \label{eq:relation_dwdl_ddeadpoints}
\end{equation}
In principle, a given variation in weight and length ($\Delta W, \Delta L$) can result from multiple distinct combinations of deadpoint adjustments. Thus, the dataset may include several input configurations that map to the same output. This ambiguity leads a neural network model to average among these cases, thereby losing valuable information.

To address the one-to-many nature of this problem, the proposed methodology adopts a different strategy: instead of modeling the direct relationship, a high-precision deep learning model is trained to learn its inverse, which is injective and thus uniquely defined.
\begin{equation}
    \texttt{machine state} + \Delta\texttt{deadpoints} \to \Delta W, \Delta L 
    \label{eq:relation_ddeadpoints_dwdl}
\end{equation}
In contrast to the previous case, the same deadpoint adjustments consistently yield the same variations in gob characteristics. 
By doing so, an optimization algorithm can be used to invert the learned model and identify the most compatible machine configuration to match the desired output.

\begin{algorithm}
\caption{Gob Deadpoint Optimization via Model Inversion}\label{alg:pseudocode}
\begin{algorithmic}
\Require Machine status: temperature $T$, master speed $MS$, firing order $FO$, number of sections N
\Require Initial deadpoints: $\{(Upper_i^0, Lower_i^0)\}_{i=0}^N$
\Require Production requirements: $\{(\Delta W_i^T, \Delta L_i^T)\}_{i=0}^N$
\Require Inversion parameters: LearningRate $\theta$, MaxSteps
\Ensure Optimized deadpoints: $\{(Upper_i^F, Lower_i^F)\}_{i=0}^N$

\State Initialize: $\{(\hat{Upper_i}, \hat{Lower_i})\}_{i=0}^N \gets \{(Upper_i^0, Lower_i^0)\}_{i=0}^N$
\State $j \gets 0$

\While{$j < \text{MaxSteps}$}
    \State Predict gob properties: 
    \State \hspace{1em} $\{(\hat{\Delta W}_i, \hat{\Delta L}_i)\}_{i=0}^N \gets \{F(T, MS, \hat{Upper}_i, \hat{Lower}_i)\}_{i=0}^N$
    \State Compute loss: 
        \State \hspace{1em} $L^j \gets \sum_{i=0}^N MSE(\Delta W_i^T, \hat{\Delta W}_i) + MSE(\Delta L_i^T, \hat{\Delta L}_i)$
    \For{$i = 0$ to $N$}
        \State Compute the gradients of the objective function:
        \State \hspace{1em} $g_i^j (\hat{Upper}_i) \gets \partial{L^j} / \partial{\hat{Upper}_i}$
        \State \hspace{1em} $g_i^j (\hat{Lower}_i) \gets \partial{L^j} / \partial{\hat{Lower}_i}$
        \State Update step:
        \State \hspace{1em} $Upper_i \gets \hat{Upper}_{i-1} - \theta \cdot g_i^j (\hat{Upper}_i)$
        \State \hspace{1em} $Lower_i \gets \hat{Lower}_{i-1} - \theta \cdot g_i^j (\hat{Lower}_i)$
    \EndFor
    \State $j \gets j + 1$
\EndWhile

\State \Return $\{(Upper_i^F, Lower_i^F)\}_{i=0}^N \gets \{(Upper_i, Lower_i)\}_{i=0}^N$
\end{algorithmic}
\end{algorithm}

\subsubsection{Regression model}
\label{subsubsec:methods_regression}

The model employed to learn the relationship presented before Equation~\ref{eq:relation_ddeadpoints_dwdl} is a fully connected neural network. This choice was motivated by the enhanced flexibility that this type of architecture offers, particularly in later stages when it becomes essential to optimize the inputs to invert the relationship.

The final architecture approximates the machine state with a set of variables, including, among the others, the temperature of the molten glass, the master speed of the plant, the feeder rotation speed and the offset between the tube and the shear mechanism. It consists of two hidden layers with 128 and 64 units respectively, separated by a BatchNormalization \cite{DBLP:journals/corr/IoffeS15} step, and it employs ReLU as activation function. Overall, the network has about 10000 trainable parameters. 

Model performance was assessed using the mean absolute error (MAE), preferred over the mean squared error (MRE) to preserve the influence of small deviations, which are prevalent in the dataset and crucial for transition stability.

To enhance regularization, the training algorithm incorporates dropout layers and employs AdamW optimization \cite{DBLP:journals/corr/abs-1711-05101}, with strong weight decay. Validation sets were carefully constructed to minimize information leakage and to exclude production batches overly similar to those in the training set, thereby ensuring a more unbiased evaluation. Given the importance of prediction accuracy for model inversion, bayesian optimization was employed to fine-tune the hyperparameters of the network. The resulting model has a relatively low parameter count, improving generalization, reducing the risk of overfitting and guaranteeing computational efficiency during the inversion. 

As discussed in Section~\ref{subsubsec:differential_approach}, the differential approach used to construct the dataset led to a high density of near-duplicate samples, especially in single-weight cycles, where inter-sectional variability is minimal. Furthermore, the data exhibits significant unbalance, with gobs concentrated in discrete weight classes (as in Figure~\ref{fig:wl_distributions}) and underrepresented in the intermediate areas between them.

To address these challenges, a duplicate removal step was integrated into the preprocessing pipeline. A $n$-dimensional histogram was constructed across combinations of input and output variables to identify overrepresented regions, from which redundant samples were selectively removed. This procedure improved data diversity and mitigated the risk of local overfitting. While alternatives such as loss reweighting or weighted sampling were considered, this histogram-based regularization proved effective in this extreme context, as the samples dropped were often more than 80\% of the training set.

Further details about evaluation strategies and performance results will be provided in Section~\ref{subsec:results_regression}.

\subsubsection{Inversion algorithm}
\label{subsubsec:methods_inversion}


As discussed in Section~\ref{subsubsec:differential_approach}, it is not feasible to directly train a model that maps the desired variations in weight and length to the corresponding adjustments in the three deadpoints, due to the non-injective nature of this relationship. To address this, once a high-precision regression model is trained to learn the opposite mapping (see Section~\ref{subsubsec:methods_regression}), the inversion is framed as a constrained optimization problem, seeking the deadpoints adjustments that would lead to the desired changes in gob properties under the current machine state.

The strategy adopted to put into practice this idea is Monte Carlo sampling, for which, maintaining fixed the set of machine operating conditions, random perturbations are applied to the three deadpoints variations of each section. The model is queried for the predicted outcome, and only those perturbations that move the prognosticated weight and length closer to the target are retained. This process is repeated iteratively until convergence. In essence, this constitutes an output-space optimization process guided by the inverse model.

While the Monte Carlo method offers robustness and flexibility, gradient-based optimization algorithms such as Stochastic Gradient Descent (SGD) or Adam can be employed for faster convergence and higher precision. These methods leverage the gradient of the loss function with respect to the input parameters, enabling more efficient navigation of the optimization landscape and the possibility to make large corrective steps from the outset. This speed is especially relevant for real-time applications, where the system must adapt to changing conditions at the rhythm imposed by the continuous operation of the machine.

The optimization is performed jointly across all sections, as the continuity of the plunger motion must be preserved. This imposes structural constraints linking adjacent Cams, which must be satisfied to ensure physically feasible transitions (see Figure~\ref{fig:cam_cycle_example}). For a machine with 8 sections, and thus 24 deadpoints, the 8 junction constraints reduce the number of free parameters to 16.

In addition to these hard constraints, soft constraints can be introduced via penalty terms in the loss function. For example, one might prefer to emphasize adjustments on the upper deadpoints over the lower ones, or enforce specific deformation patterns on the Cam profile to preserve mechanical safety or process efficiency.

\begin{figure}
    \centering
    \includegraphics[width=\linewidth]{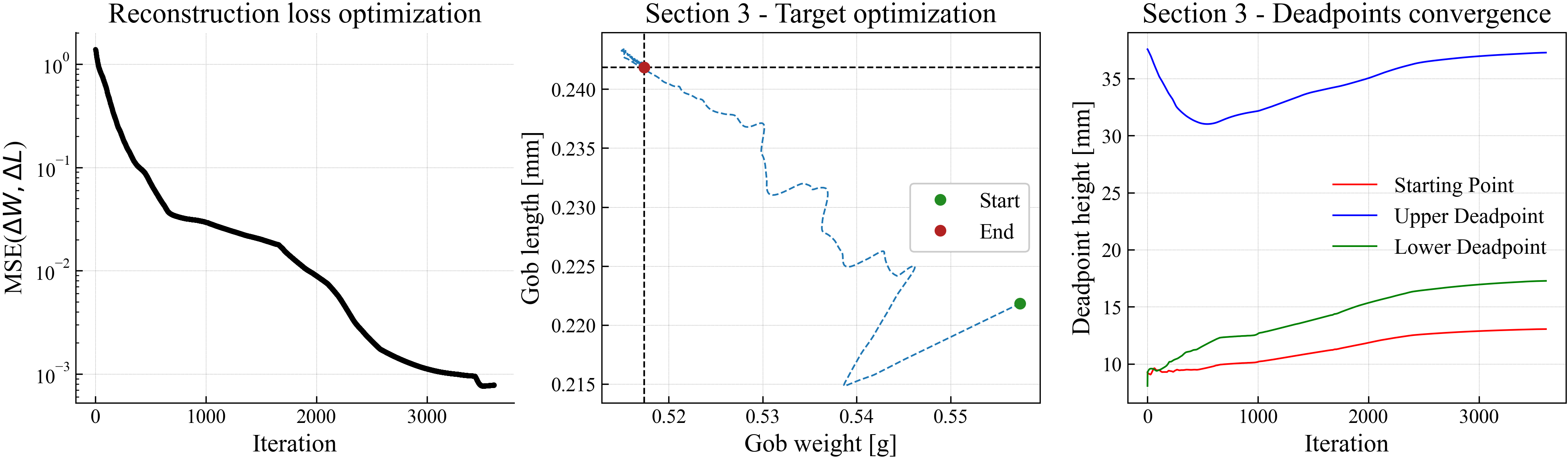}
    \caption{Simulated inversion on a sample cycle. The image reports the convergence progress observed on one of the eight sections for both the targets and the inputs. The transformation requested is random in the range [-40g, +40g] for the weight and [-20mm, +20mm] for the length.}
    \label{fig:inversion_example}
\end{figure}

Figure~\ref{fig:inversion_example} illustrates an example of the inversion algorithm applied to a few sections, showing the evolution of both the deadpoints and the resulting gob characteristics throughout the process. The image highlights that the optimization trajectory is not linear towards the target values, but instead exhibits oscillations due to the continuity constraints imposed by neighboring Cam profiles. Nevertheless, convergence toward the desired output is consistently achieved.

Operationally speaking, once the feeder tube height has been set, the inversion algorithm is deployed in real-time, acquiring the current state of the machine and computing the deadpoints adjustments required to meet the target gob specifications.

It is worth noticing that the non-injectivity of the original problem does not entirely vanish in this formulation. Depending on the initial condition of the optimizer, different configurations may converge to equally valid solutions, all producing the desired gob characteristics. While this is acceptable from a production standpoint, future work could explore ways to guide the optimizer toward more desirable or interpretable configurations.

\section{Results and Discussion}
\label{sec:results_discussion}

This section presents the results obtained with the methodology proposed in Section~\ref{subsec:methods}, and discusses their implications and limitations when applied to the field of hollow glass manufacturing.

The discussion begins by evaluating the performance of the regression model, with particular focus on the choice of validation metrics and the sampling strategy adopted for the validation sets.

Next, the power of the inversion algorithm is assessed through its application on some real-world scenarios, where it attempts to reconstruct the deadpoints adjustments that led to specific gob characteristics.

Although real-time testing on the production line was not feasible, the evaluation on historical data still provides meaningful insights into the algorithm’s behavior and its potential impact in a real-world setting.

\subsection{Regression results}
\label{subsec:results_regression}

Given the objective of obtaining unbiased and realistic estimates of the model’s generalization capabilities, special care was taken to ensure that the validation data reflects the diversity and imbalance of the actual production scenarios. Providing an unbiased yet comprehensive evaluation of the regression model's performance is challenging primarily due to the nature of the data, as they originate from historical records of relatively stable production cycles, during which the same operational patterns were repeated over extended periods. This lead to a dataset a high incidence of duplicate or near-duplicate input-target combinations, introducing a significant risk of information leakage in case of an unwise choice of training and validation subsets.

A naive random split of the data would likely result in the model being evaluated on samples that closely resemble those seen during training, thereby inflating performance metrics. This phenomenon was confirmed by several experiments, in which model accuracy on randomly selected validation sets closely mirrored that of the training set. To obtain a more rigorous and objective evaluation, a temporal data split was adopted: the earliest months of data were used for training, while the most recent (typically one-quarter of the total) were reserved for validation. This approach not only minimizes the risk of overlap between training and test samples, but also better simulates real-world deployment scenarios, in which the model would be periodically retrained as new data becomes available. However, as will be shown, this strategy may require the model to extrapolate beyond the domain on which it was originally trained.

\begin{figure}[!ht]
    \centering
    \includegraphics[width=\linewidth]{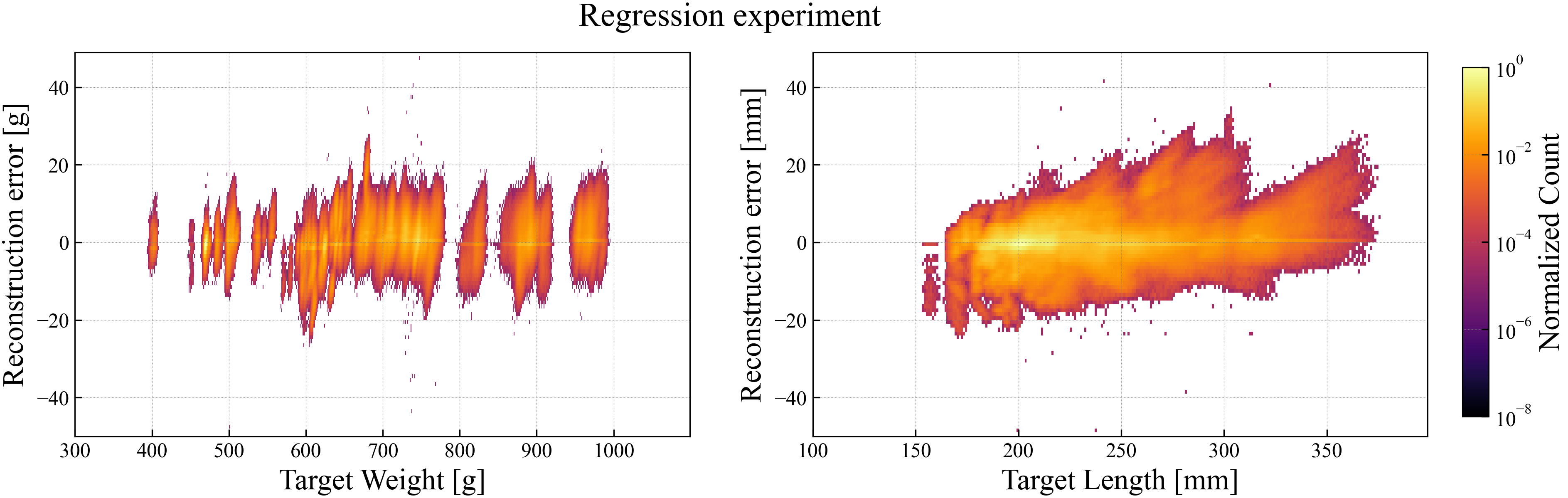}
    \caption{Results of a regression experiment over line A-1. The validation set is constituted by the productions relative to the last 4 months of the dataset. The image depicts the distribution of the reconstruction errors of the model, highlighting the Weight/Length classes.}
    \label{fig:regression_results21}
\end{figure}

Figure~\ref{fig:regression_results21} summarizes the outcomes of a regression experiment conducted on line A-1. The dataset for this line spans approximately 16 months of measurements, with the final four months allocated for validation. To highlight the pronounced class imbalance in the data, both predictions and targets are visualized using two-dimensional histograms, which offer clearer insight than traditional scatter plots.

Numerical results are summarized in Table~\ref{tab:regression_results21} that reports the Mean Absolute Errors (MAE) not computed on the raw variations ($\Delta W$ and $\Delta L$), but rather on the final gob characteristics ($W_{start} + \Delta W$ and $L_{start} + \Delta L$). This adjustment accounts for the fact that higher weight classes inherently tolerate greater variability, while smaller classes demand higher precision.

As detailed in Section~\ref{subsubsec:methods_regression}, the histogram-based duplicates removal step helped in mitigating the effects of data unbalance. This process led to the elimination of 60\%–80\% of the samples, though the remaining data still exhibited a distribution centered irregularly around zero. The effectiveness of this procedure fined parameters, such as the granularity of the n-dimensional bins (i.e., the desired sensitivity) and the choice of input variables. Also, the extent of sample removal has a direct impact on final performance metrics. The values reported in Table~\ref{tab:regression_results21} are calculated on the reduced dataset and are thus considered to provide a more unbiased, yet conservative, estimate of the model’s accuracy, with reduced sensitivity to the effects of long, repetitive productions.

\begin{table}[!ht]
\centering
\begin{tabular}{@{}r|cc}
\toprule
    & \textbf{MAE [Weight]} & \textbf{MAE [Length]} \\
    \midrule
    \textbf{Train}      & 3.4 g & 2.3 mm \\
    \textbf{Validation} & 3.9 g & 3.4 mm \\
    \bottomrule
\end{tabular}
\caption{Metrics relative to the regression experiment on line A-1 described in Figure~\ref{fig:regression_results21}.}
\label{tab:regression_results21}
\end{table}

For completeness, analogous experiments were conducted on the other available production lines (A-2, A-3, and B-1), following the same validation strategy by isolating the most recent months of data. The corresponding performance metrics are summarized in Table~\ref{tab:regression_results_otherlines}.

An extended evaluation of the regression performance is presented in \ref{app:extended_eval_regr}, while a more detailed discussion about the results achieved will be provided in Section~\ref{subsec:discussion}.

\begin{table}[!ht]
\centering
\begin{tabular}{@{}r|cc|cc@{}}
\toprule
    \multicolumn{1}{l|}{} & \multicolumn{2}{c|}{\textit{Training Set - MAE}} & \multicolumn{2}{c}{\textit{Validation Set - MAE}} \\
     & \textbf{Weight} & \textbf{Length} & \textbf{Weight} & \textbf{Length} \\
    \textbf{Line ID} & & & & \\
    \midrule
    A-2 & 2.6 g & 2.9 mm & 6.6 g & 6.4 mm \\
    A-3 & 4.5 g & 3.5 mm & 7.8 g & 6.1 mm \\
    B-1 & 4.2 g & 1.5 mm & 8.9 g & 6.0 mm \\ 
    \bottomrule
\end{tabular}
\caption{Metrics obtained from regression experiments performed on the lines A-2, A-3 and B-1. They all consist of around 12 to 15 months of measurements, the last 25\% of which constitute the validation set.}
\label{tab:regression_results_otherlines}
\end{table}

\subsection{Inversion results}
\label{subsec:results_inversion}

The ideal method for validating the inversion algorithm would involve real-time deployment, as this would allow for a direct assessment of the algorithm’s accuracy in implementing various transformations (of the weight and length of the gobs produced) to an arbitrary initial production cycle. However, the primary challenge in this validation approach lies in the non-injectivity of the mapping being inverted (Equation~\ref{eq:relation_dwdl_ddeadpoints}), which implies that a unique solution is not guaranteed in all scenarios. As a result, the algorithm may converge to multiple Cam configurations that are equally valid from a production standpoint, yet not necessarily identical to those historically implemented.

This ambiguity stems from the model’s training strategy, which, as described in Section~\ref{subsubsec:differential_approach}, relies on individual Cam transformations, effectively treating each section as an independent entity. In contrast, the inversion procedure must consider the cycle as a whole, where the deadpoint corrections are inevitably shaped by the boundary conditions imposed by adjacent Cams. Consequently, the convergence outcome reflects not only the desired transformation but also the global interdependencies within the cycle.

In light of these constraints, validation of the technique is performed using the available historical dataset. Specifically, the model’s output is evaluated by quantifying the degree of convergence across previously implemented production cycles and comparing the algorithm’s predicted corrections to those that were actually applied in practice.

The evaluation was conducted using data from line A-1, by selecting a representative subset of 1,000 cycles from the validation set presented in Figure~\ref{fig:regression_results21}. Given that each optimization run typically requires approximately ten seconds, depending on the magnitude of the requested transformation, it would be computationally inefficient to apply the inversion algorithm across the entire dataset. To ensure fair representation of the diverse production scenarios, an $n$-dimensional histogram was constructed based on the input variables used by the model. This approach, analogous to the duplicates removal process described in Section~\ref{subsubsec:methods_regression}, was employed to guide the sampling procedure and avoid selecting cycles with overly similar feature combinations.

The algorithm was specifically optimized to achieve such a reduced convergence time. In the industrial plants analyzed, the master speed, i.e. the machine’s operating frequency, typically stabilizes at 6–8 cuts per section per minute. This corresponds to an average cycle time of approximately 10–12 seconds. For this setup, a functional algorithm must therefore deliver results within an optimal timeframe, incurring at most the loss of 2–3 production cycles.

\begin{figure}
    \centering
    \includegraphics[width=\linewidth]{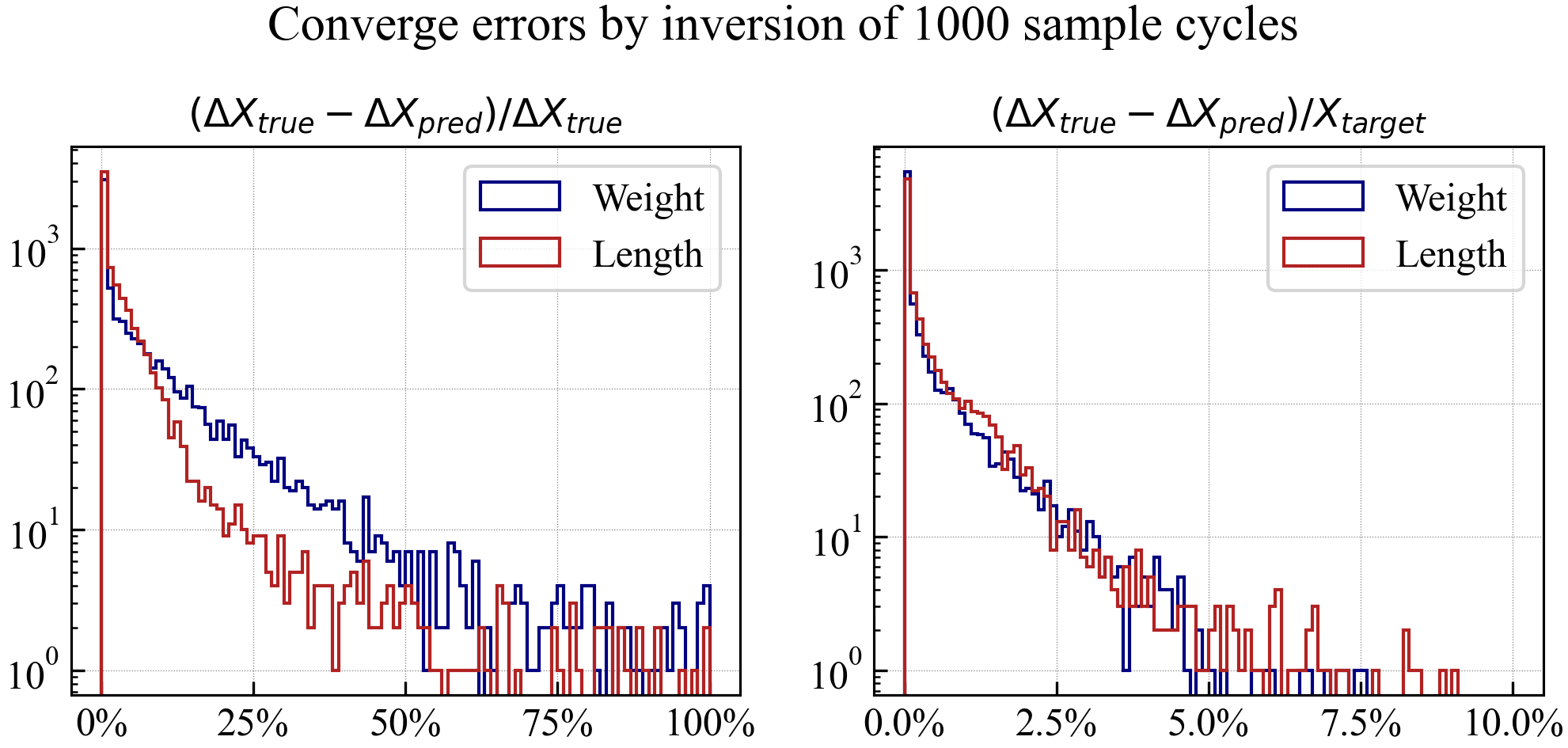}
    \caption{Relative errors with respect to the inversion targets ($\Delta W$ and $\Delta L$), computed over 1,000 cycles extracted from the validation set of line A-1. The left panel shows the error distribution relative to the requested transformation, while the right panel shows the error relative to the final gob characteristics.}
    \label{fig:convergence_errors_inversion}
\end{figure}

\begin{figure}
    \centering
    \includegraphics[width=0.6\linewidth]{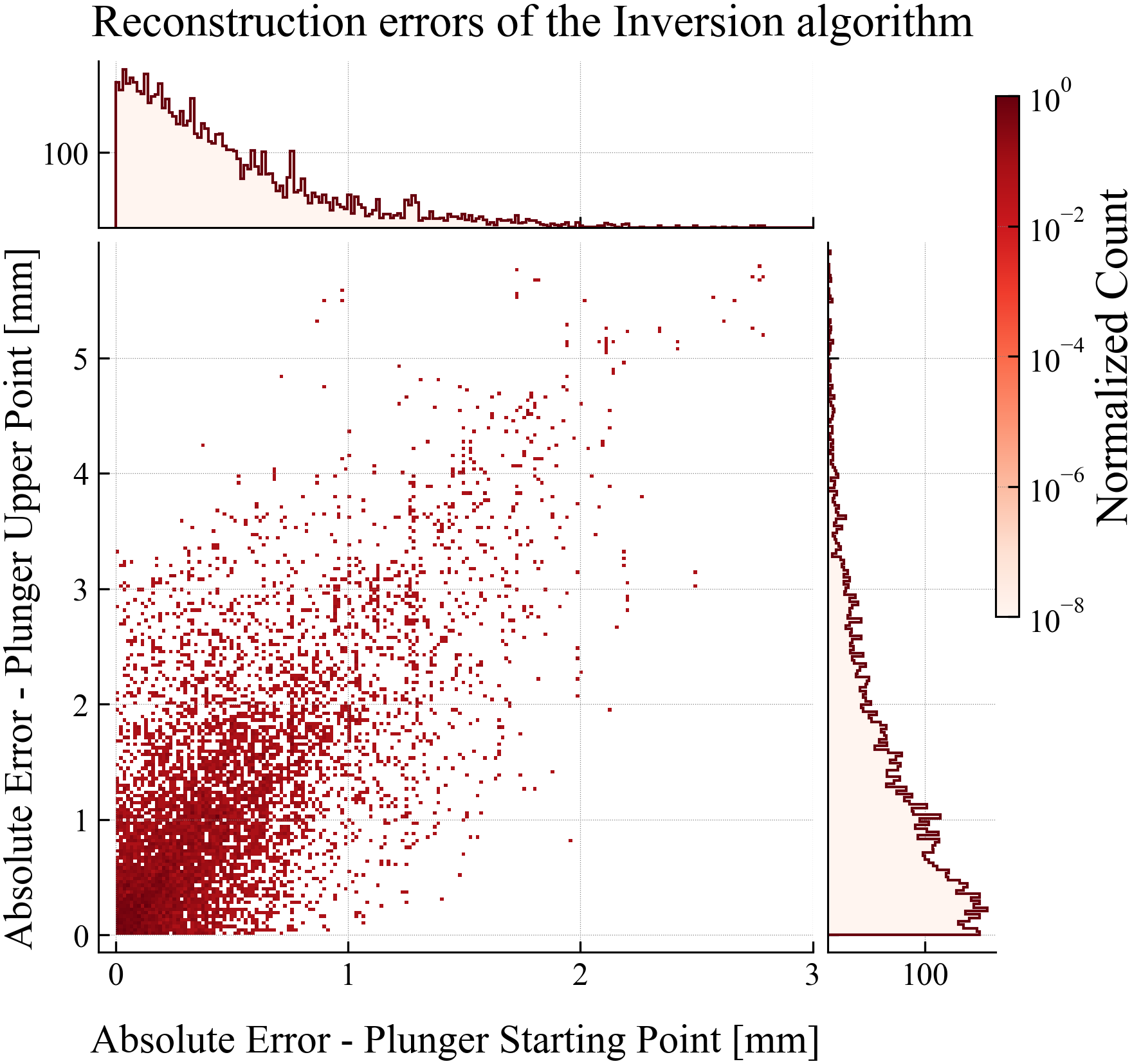}
    \caption{Reconstruction errors of the deadpoint corrections predicted by the inversion algorithm, evaluated over 1,000 cycles from the historical dataset of line A-1.}
    \label{fig:reconstruction_errors_inversion}
\end{figure}

Figures~\ref{fig:convergence_errors_inversion} and \ref{fig:reconstruction_errors_inversion} present the results obtained by applying the inversion algorithm to a subset of 1,000 cycles, sampled from the most recent months of data from line A-1. The former illustrates the errors on the inversion targets (i.e., the requested variations in Weight and Length), thereby demonstrating the degree of convergence achieved by the algorithm. In contrast, the latter displays a two-dimensional histogram of the discrepancies between the deadpoint corrections proposed by the algorithm and those that were actually applied in practice to implement the transformation.

\subsection{Discussion}
\label{subsec:discussion}

As shown by the metrics in Table~\ref{tab:regression_results_otherlines} and, specifically for line A-1, by Table~\ref{tab:regression_results21} and Figure~\ref{fig:regression_results21}, the regression model proves capable of meeting the requirements necessary for integration within the inversion system, explained in Section~\ref{subsubsec:methods_inversion}. It demonstrates high accuracy in reconstructing known production transformations and exhibits solid generalization performance on validation sets, constructed to stress exactly these characteristics.

Nonetheless, the model exhibits limitations when extrapolating to regions of the input space that differ significantly from those seen during training. This poses a potential challenge when deploying the framework on entirely new production setups, particularly if they fall outside the historical operating envelope. However, as more diverse and representative data become available over time, the model’s effective working domain is expected to expand, progressively improving its robustness and reliability in broader scenarios.

A noteworthy case, among those analyzed, is represented by line A-2, which exhibits comparatively lower performance relative to the other lines. The underlying reason becomes evident when examining the spectrum depicted in Figure~\ref{fig:wl_distributions}, where it can be observed that the working point for this line is concentrated around significantly lower values, in terms of weight and length classes. As a result, although the absolute prediction errors of the model are comparable to those seen in other cases (Table~\ref{tab:regression_results_otherlines}), the accuracy, defined using fixed relative thresholds, appears disproportionately affected due to the increased sensitivity.

Numerous additional tests were conducted to refine the neural network's architecture and parameters, as well as to ensure the stability and generalizability of its predictions. However, these results fall outside the primary scope of this paper and are therefore not included.

The validation tests proposed in Section~\ref{subsec:results_inversion}, which pertain to the inversion phase of the proposed framework, focus primarily on two aspects: the optimizer’s ability to converge reliably and its accuracy in reconstructing the actual deadpoints historically applied to implement specific transformations.

As depicted in Figure~\ref{fig:convergence_errors_inversion} (on the right), the algorithm successfully converges to the specified Weight and Length targets in the vast majority of the 1,000 cycles evaluated, consistently meeting the defined accuracy thresholds. The left panel, which depicts relative errors with respect to the requested variation (rather than the final target), exhibits a few more outlier cases. This discrepancy is primarily attributed to very small requested changes, typical of single-weight production scenarios, that amplify relative error due to small denominators. Additionally, the use of a logarithmic scale emphasizes these extreme cases, which nonetheless represent a small fraction of the total.

Figure~\ref{fig:reconstruction_errors_inversion} shows the distribution of reconstruction errors for the deadpoints, confirming that the algorithm is generally able to accurately reproduce the actual Cam settings used during production. The histogram indicates that, in most cases, absolute errors for the upper deadpoint are below 1 mm, while starting deadpoint errors typically remain well under 0.5 mm. According to domain experts, such minor variations in deadpoints, less than 1 mm, usually result in negligible differences in gob weight and length (on the order of a few grams or millimeters), thereby falling within acceptable production tolerances.

Overall, the challenges introduced by the non-injectivity of the mapping being inverted (Equation~\ref{eq:relation_dwdl_ddeadpoints}) appear to be effectively mitigated by the continuity constraints imposed by adjacent cams in the production cycle, at least in scenarios involving isolated transformations of a single Cam.

\subsubsection{Interpretation}
\label{subsubsec:interpretation}

The stability of the optimization step enables the inversion algorithm to offer valuable insights into machine behavior, particularly regarding the typical deadpoints variations required to achieve specific gob characteristics.

\begin{figure}
    \centering
    \includegraphics[width=.9\linewidth]{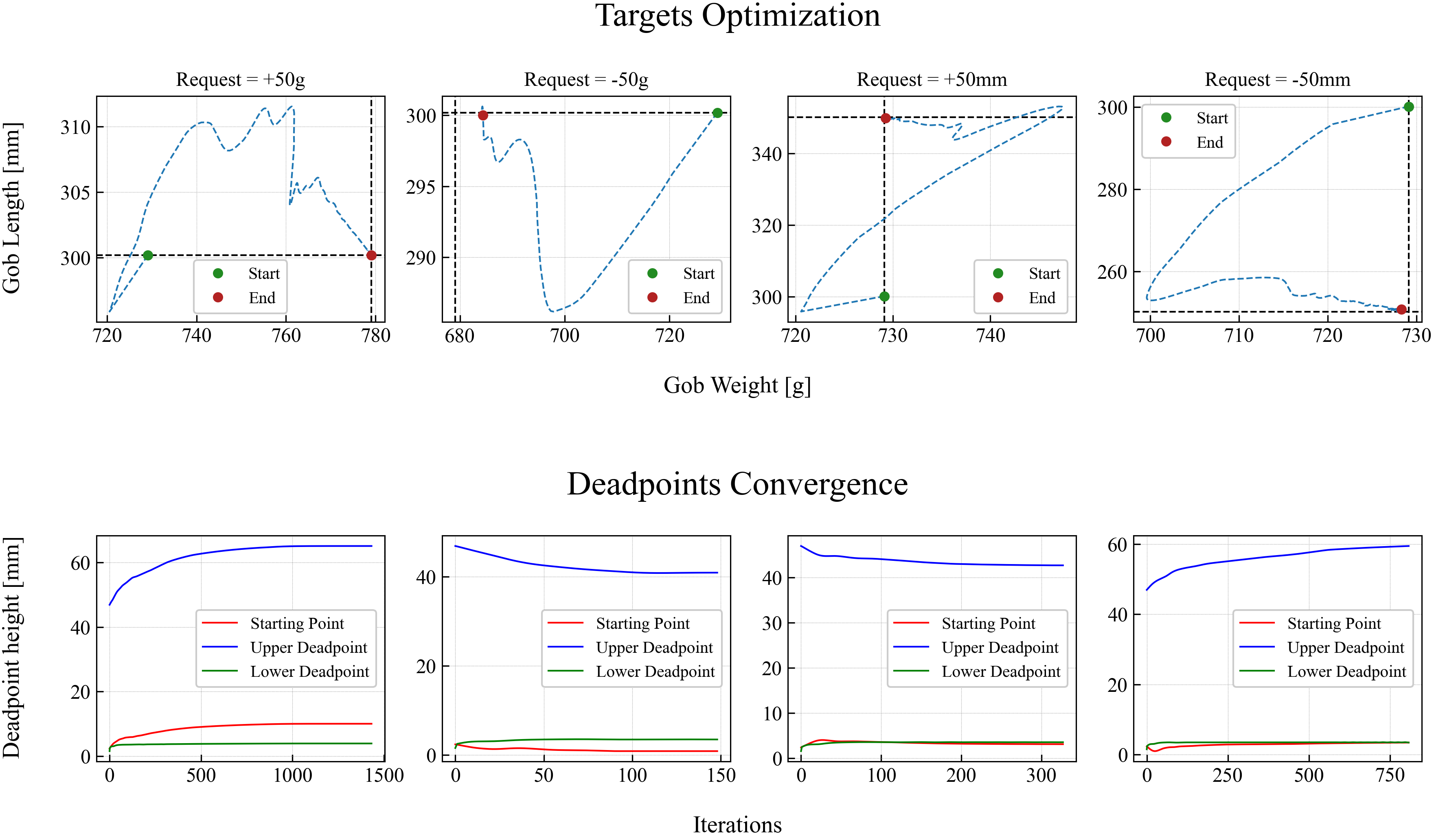}
    \caption{"Playing" with the optimization algorithm: the user requests, reported in the subplots' titles, is applied on a random section of an example cycle whose other gobs remain unchanged. The first row reports the progress of the transforming section characteristics, while in the second row the consequent changes in the deadpoints is shown.}
    \label{fig:optimization_interpretation}
\end{figure}

Figure~\ref{fig:optimization_interpretation} depicts the potential changes to cams profiles in response to simulated, realistic user requests applied to a single gob during an initial production cycle. The plotted trajectories represent the evolution of the target characteristics throughout the optimization process, during which the algorithm adjusts all deadpoints concurrently to align the changing section with the desired transformation. The observed fluctuations in Weight-Length values primarily result from boundary conditions imposed by adjacent cams (whose gob characteristics are set to remain unchanged), which constrain the optimization of the targeted cam.

By imposing targeted and isolated modification requests, several key behavioral patterns can be inferred. These trends have been validated by experienced personnel in the field, thus reinforcing the interpretability and practical reliability of the algorithm.

Furthermore, instances in which the algorithm fails to converge can be interpreted as indicators of physically unfeasible transformation requests (because of the mechanical limitations of the plunger system) since the model is unable to identify a viable configuration of deadpoints that satisfies the desired gob characteristics.

\subsubsection{Stability}
\label{subsubsec:stability}

The stability of the algorithm can be assessed by analyzing its response to a series of progressively increasing transformation requests, thereby evaluating the consistency and robustness of its responses.

The experiment involves selecting a few representative production cycles from the validation set of line A-1. For each cycle, a section is randomly chosen and subjected to controlled transformations within the ranges of [-50g,+50g] for weight and [-50mm,+50mm] for length, while the features of all other sections are constrained to remain unchanged.

\begin{figure}
    \centering
    \includegraphics[width=.9\linewidth]{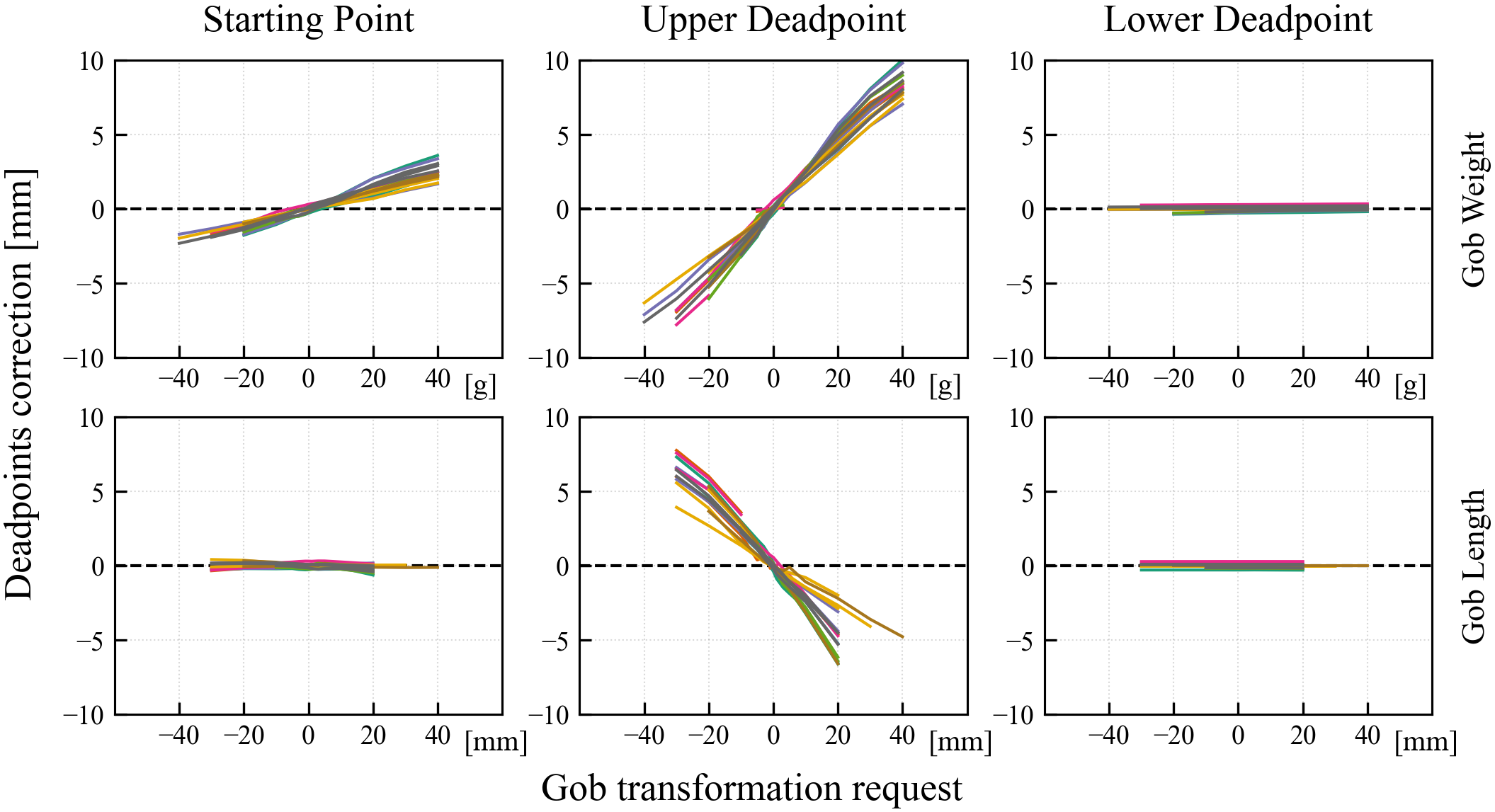}
    \caption{Consistency of the algorithm's response across different transformation scenarios. deadpoints' behavior is monitored under gradually varying Weight-Length changes applied by the user.}
    \label{fig:inversion_stability}
\end{figure}

Simulation results are shown in Figure~\ref{fig:inversion_stability}. As can be observed, the algorithm consistently produces stable corrections across a wide range of transformation requests. These results further support the findings discussed in Section~\ref{subsubsec:interpretation} regarding typical operational behaviors.

During such isolated transformations, the adjusted starting point may induce slight perturbations in the lower deadpoint of the adjacent cam on the left, which in turn compensates to preserve the physical characteristics of its gob. This highlights how the algorithm’s modifications on a single Cam can propagate through the entire cycle, with neighboring sections progressively absorbing part of the change.

Some of the curves in the figure appear truncated, indicating failed convergence by the algorithm for certain transformation requests. These cases typically arise when the required corrections exceed the mechanical limits of the plunger, such as when the only possible solution would involve setting negative deadpoints. The variation in slope among the curves likely reflects differences in the machine states of the respective cycles, potentially associated with variations in glass temperature and viscosity.

\begin{figure}
    \centering
    \begin{subfigure}[b]{0.45\textwidth}
        \includegraphics[width=\textwidth]{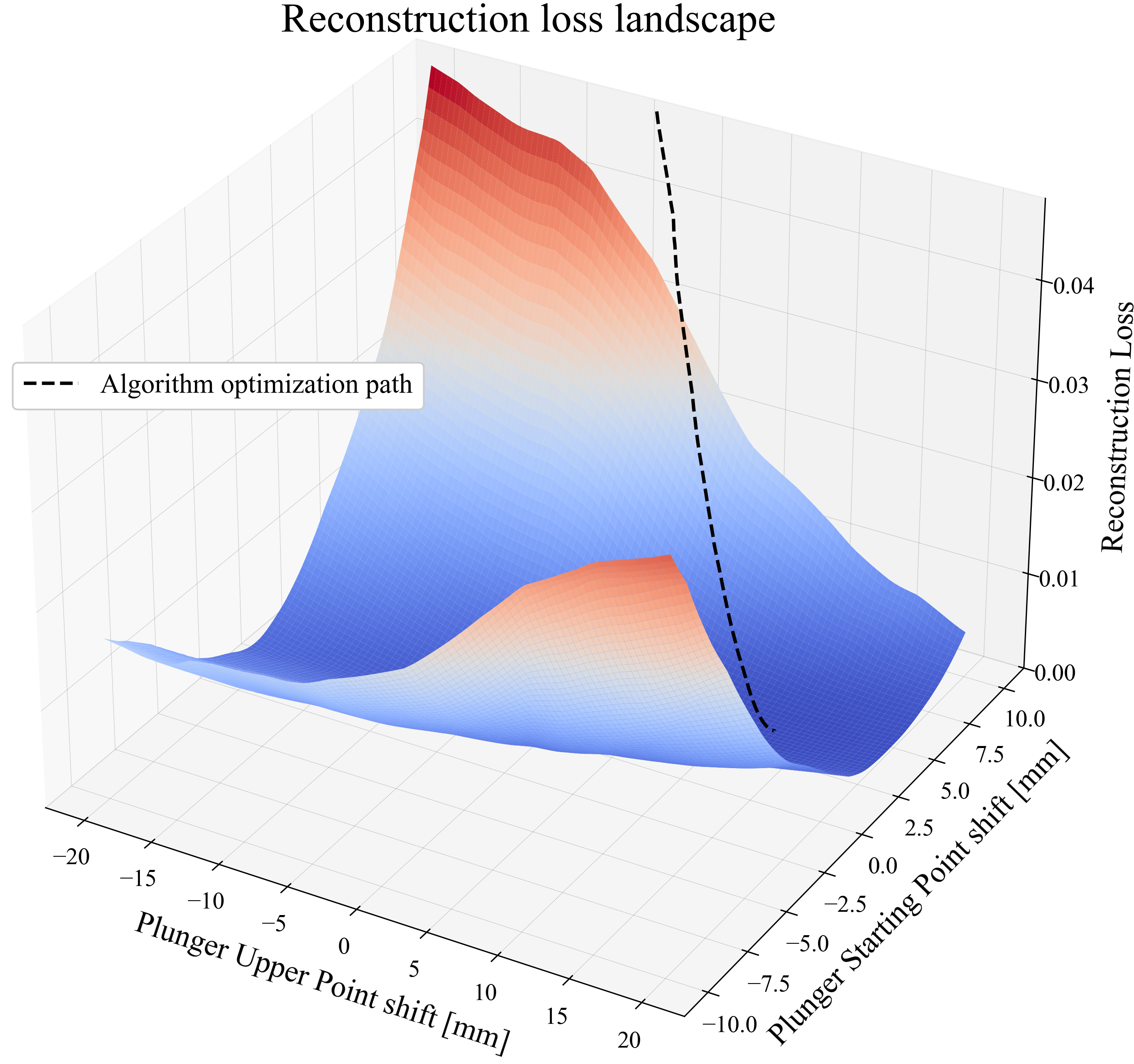}
        \caption{Surface plot}
        \label{fig:loss_landscape_surface}
    \end{subfigure}
    \hfill
    \begin{subfigure}[b]{0.45\textwidth}
        \includegraphics[width=\textwidth]{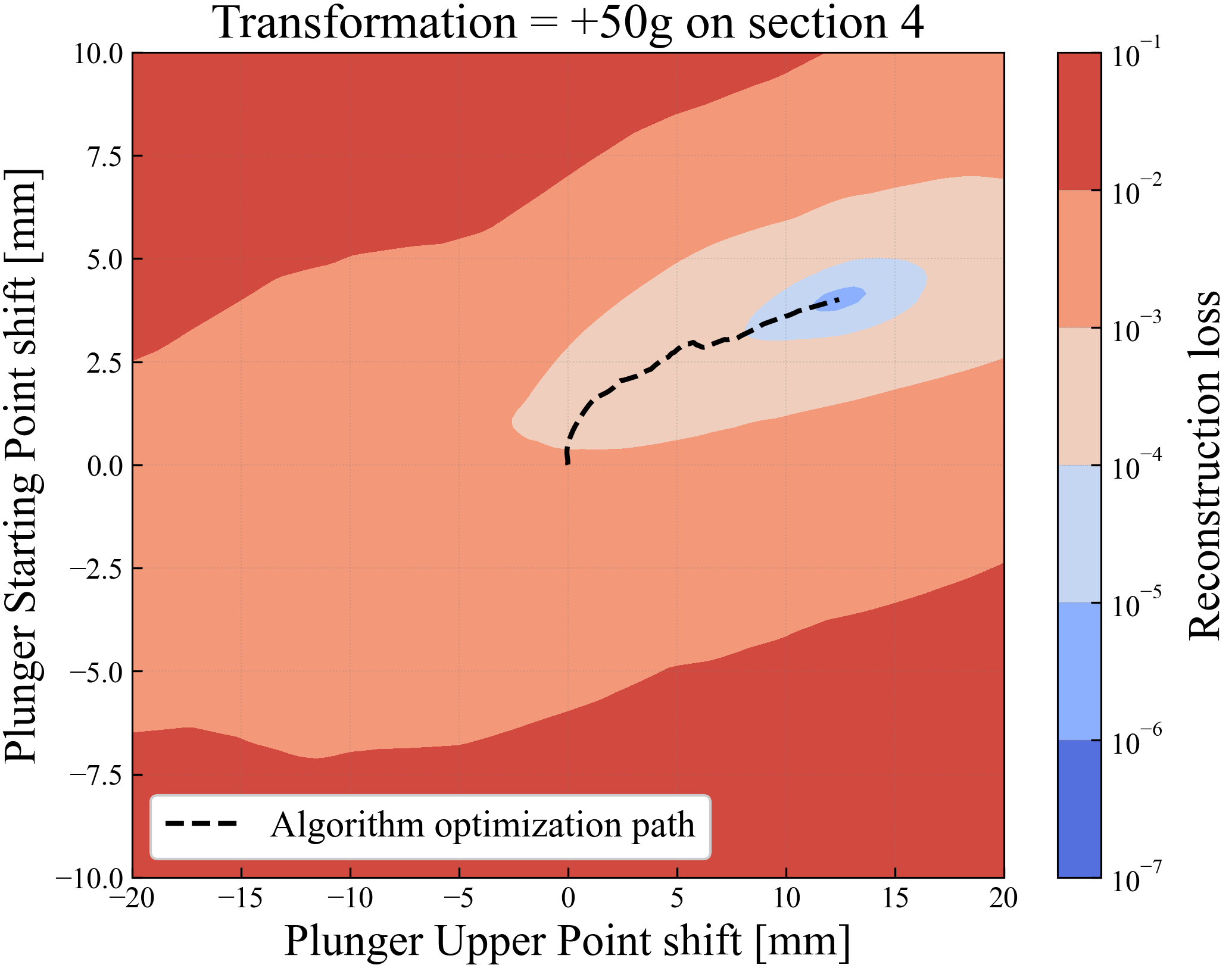}
        \caption{Contour lines}
        \label{fig:loss_landscape_contourf}
    \end{subfigure}
    \caption{Given an example transformation (+50 g) over a Cam, the Figure provides a bi- and tri-dimensional representation of the deadpoints' reconstruction loss landscape. The optimization algorithm (black dashed line) will move across this irregular space to find the minimum that represents the optimal final configuration.}
    \label{fig:loss_landscape}
\end{figure}

Visualizing the loss landscape provides valuable insights into the behavior of the inversion algorithm. This can be achieved by selecting a random cycle and requesting a change (in this case, an increase of +50g of weight) on one section’s characteristics, while keeping the others constant. Figure~\ref{fig:loss_landscape} illustrates the surface and the contour lines of the loss function corresponding to the regression model, as it responds to various adjustments in the starting and upper deadpoints of the involved cam. The curves represent the actual high-dimensional loss surface over which the optimizer navigates in its effort to locate the minimum of the landscape, as depicted by the dashed line in the plot.

As shown in the figure, the surface is relatively smooth, and the optimizer encounters little difficulty in steadily converging toward the function’s minimum. This indicates that the algorithm is capable of identifying an optimal cam configuration that satisfies the transformation requirements effectively. 

Furthermore, in this case, the structure of the contour lines appears to suggest the presence of a global minimum, corresponding to a unique target configuration (aside from minor fluctuations). This behavior is most likely attributable to the boundary conditions imposed by adjacent Cams, which constrain the solution space and, in effect, reduce the non-injectivity of the mapping described in Equation~\ref{eq:relation_dwdl_ddeadpoints}. As a result, the inversion process becomes more stable and deterministic within this local operating context.

\subsubsection{Existence of multiple solutions}
\label{subsubsec:multiple_solutions}

The primary challenge in demonstrating the existence of multiple solutions lies in the high number of degrees of freedom typically involved in an optimization process.

\begin{figure}[!ht]
    \centering
    \includegraphics[width=\linewidth]{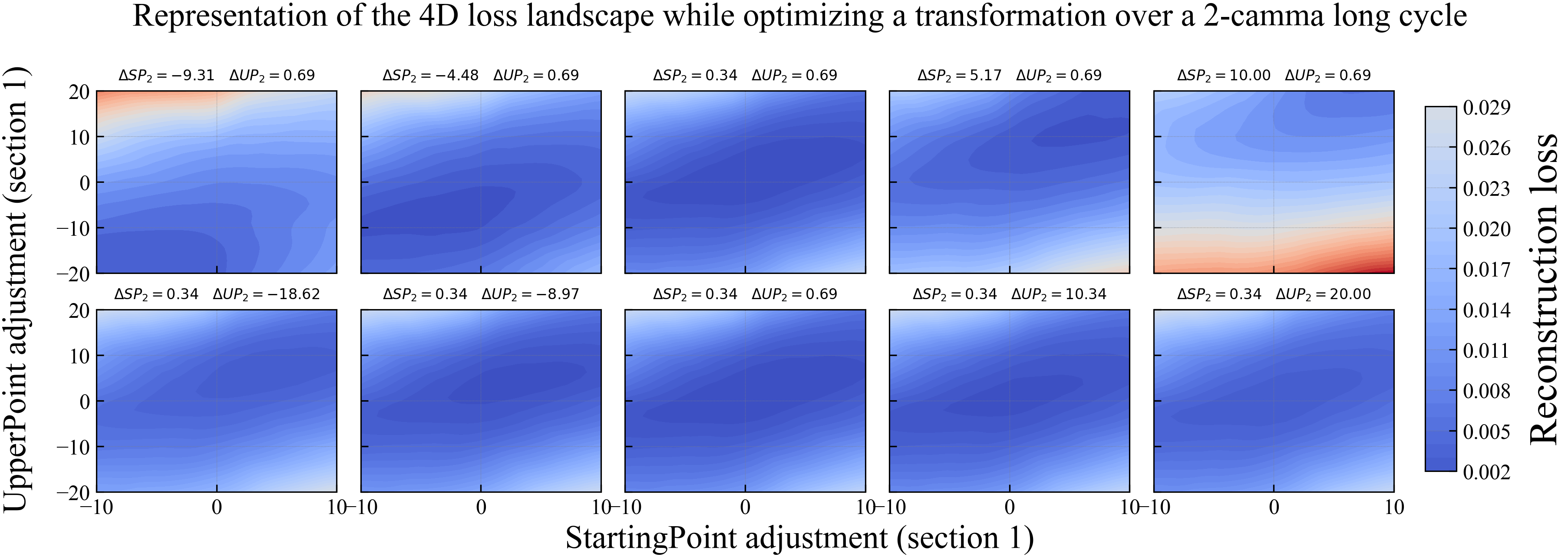}
    \caption{Loss landscape representation for a simulated transformation on a 2-Cam cycle. Each plot represents the contour lines of the model loss for different combinations of deadpoints for the first Cam, at different fixed configurations of the second Cam. Each local minimum represents a valid gob configuration: most of them, however, cannot be reached by the optimizer without forcing large steps in a specific direction. The 4D evaluation grid of the loss landscape has a size of $30^4$.}
    \label{fig:multiple_minima_landscape}
\end{figure}

To illustrate their existence more clearly, a simplified transformation experiment has been simulated, involving a cycle with only two Cams. Reducing the system in this way reduces the number of degrees of freedom to four (six deadpoints minus two boundary conditions), making the problem easier to visualize. A small transformation was applied to both Cams: a small increase of +10 grams to the first section and a decrease of -10 grams on the second, while maintaining their length constant.
In continuity with the representation introduced in Section~\ref{subsubsec:stability}, Figure~\ref{fig:multiple_minima_landscape} depicts the optimization loss landscape corresponding to the first Cam for various combinations of adjustments of the second one. The visualization explicitly shows that, depending on the final configuration of the second Cam, the constrained loss landscape of the first one gradually shifts, causing the gradient descent algorithm to converge to different minima.
The optimizer presented in this work always starts from a null variation $(\Delta SP_{i}=0,\Delta UP_{i}=0)$ (the center of the contour plot) and therefore converges to the closest minimum, assuming that the loss landscape does not present local irregularities. Table~\ref{tab:multiple_minima_solution} reports the actual solution found by the proposed setup.

From an industrial perspective, this behavior is highly desirable: although each minimum yields a valid gob configuration, the inherently slow dynamics of molten glass make large, abrupt changes in deadpoint positions both inefficient and potentially harmful, as they can induce mechanical stress or exceed the allowed operating range. Consequently, the optimal solution is the one that minimizes global setup adjustments, which is precisely the behavior ensured by this optimization algorithm.

\begin{table}[!ht]
    \centering
    \begin{tabular}{cccc}
    \toprule
    $\Delta SP_1$ & $\Delta UP_1$ & $\Delta SP_2$ & $\Delta UP_2$ \\
    \midrule
    +0.31 mm & +1.90 mm & -0.65 mm & -2.34 mm \\
    \bottomrule
    \end{tabular}
    \caption{Solution found for the 2-Cam cycle transformation. The point found is compatible with the combinations provided in Figure~\ref{fig:multiple_minima_landscape} and, because of the geometry of the problem, it represents the closest deadpoints variation which generates the gob with the given characteristics.}
    \label{tab:multiple_minima_solution}
\end{table}

\subsubsection{Remarks on Data-Driven Approaches}
Data-driven methods, such as the proposed algorithm, inherently rely on the quality and coverage of the available data. As discussed in Section~\ref{subsec:discussion}, the regression phase naturally exhibits limited generalization in regions of the Weight-Length spectrum that are underrepresented in the training data, particularly beyond the boundaries of the dataset. To address this, it is crucial to compile a diverse and representative dataset, enabling the model to encounter a broad range of input scenarios and learn the underlying mapping described in Equation~\ref{eq:relation_ddeadpoints_dwdl} more effectively. In cases where data availability is limited, techniques such as regularization and pre-training with measurements from various production setups can help improve robustness. However, it remains challenging to guarantee the system's performance on previously unseen production configurations with substantially different characteristics, reflecting an intrinsic aspect of data-driven modeling. An optimized training process on a diversified dataset typically results in a more accurate model and a smoother loss landscape, which also facilitates the inversion process.

Regarding the inversion procedure, certain challenges are fundamental to the nature of data-driven approaches. For instance, optimizer failures to converge - such as when it cannot find a suitable set of deadpoints — often occur when the requested transformations exceed the machine's physical constraints, representing inherent feasibility issues rooted in the problem formulation itself. Decomposing such transformations into smaller, manageable steps can help maintain operational validity.

More fundamentally, scenarios where the algorithm converges successfully but produces an invalid solution highlight a characteristic challenge of inverse modeling based on data-driven methods. These cases are particularly challenging to detect before deployment, as the inverse transformation (Equation~\ref{eq:relation_dwdl_ddeadpoints}) is non-injective, allowing for multiple valid solutions—some of which may not align with historical data. Such phenomena often originate from an incomplete or imbalanced training dataset, which affects the learned loss landscape in certain regions of the input space. Addressing this involves expanding the dataset and improving the model's representation of the operating domain, a natural and ongoing aspect of data-driven development.

Finally, the overall error in the algorithm's output stems from multiple sources: the intrinsic reconstruction error of the model, the convergence behavior (which is generally robust), and the discrepancy between simulated predictions and real-world physical behavior. These factors are typical of data-driven approaches and are best evaluated through targeted real-world testing.

\subsubsection{Impact on production time}

\begin{figure}[!ht]
    \centering
    \includegraphics[width=\linewidth]{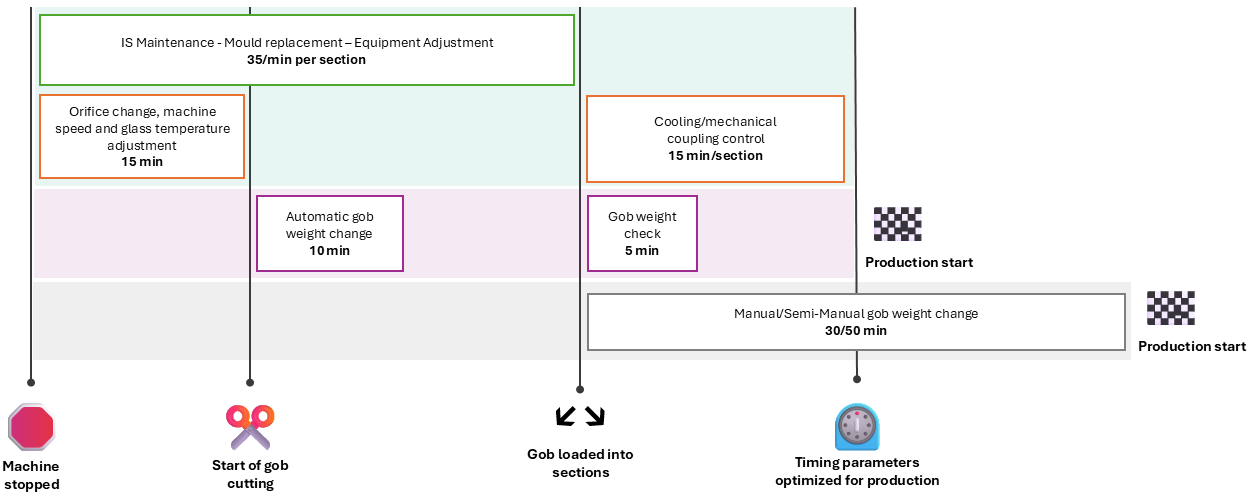}
    \caption{Schematic timeline for a new production setup. The green shadow represents irreducible steps to set up the production line. The grey box represents the traditional setup procedure, while the purple boxes illustrate the proposed automatic configuration algorithm.}
    \label{fig:setup_timeline}
\end{figure}

A major benefit of the proposed inversion framework is the reduction in setup time during production change-overs. Figure~\ref{fig:setup_timeline} represents the schematic representation of the current setup process. The green-shaded area represents common steps that are needed before starting a new production, which are irreducible. The grey box represents the traditional setup procedure, while the purple boxes illustrate the proposed automatic configuration algorithm. The key difference lies in when and how the gob parameters are evaluated and adjusted.  

In the current manual process, operators must wait until the machine reaches full mechanical and thermal coupling—after gob loading—so that the produced bottles can be inspected and their characteristics assessed. Based on this feedback, they iteratively adjust the plunger dead-points by hand, often repeating the cycle several times until the desired gob weight and shape are achieved. This procedure makes the setup phase lengthy, highly dependent on operator experience, and potentially hazardous due to the manual handling of hot glass.

In contrast, the proposed method leverages real-time measurements of gob weight and length acquired by an infrared sensor immediately after maintenance and before the gobs are distributed to the forming sections. By analyzing these measurements, the algorithm can estimate the necessary dead-point corrections automatically, allowing production to start as soon as the gobs are stable—without waiting for full thermal and mechanical equilibrium or performing manual iterations.

Because the problem is non-injective, multiple parameter configurations can yield gobs with identical physical characteristics; both the manual and automatic setups thus lead to valid operating conditions for initiating production.
However, the automatic configuration selects among these equivalent solutions the one that is closest to the previous machine state, thereby ensuring a smooth transition, minimizing actuator stress, and significantly reducing setup time—from approximately 30–50 minutes to about 10 minutes (a 65–80 
This early-correction stage shortens downtime, reduces manual intervention, and allows operators to supervise the process safely from a control room.

\subsection{Future Work}

While the current results demonstrate the effectiveness and robustness of the proposed approach, several directions remain open for further investigation.

A more dedicated and in-depth analysis of the network's loss landscape could potentially offer additional insights into the optimization dynamics and interpretability of the process. Additionally, a broader and more varied dataset would likely enhance the model's generalization capacity and help clarify its failure modes.

In the current machine configuration, the raw acquisitions of the sensors consist of infrared images (potentially three-dimensional in the case of more advanced systems) of the falling gobs. These images are subsequently processed to extract weight and length measurements, which form the core of the dataset used in this study. A promising direction for future research involves integrating additional features into the model, such as gob shape, volume, or the spatial distribution of molten glass. Information regarding the physical state of the glass, such as viscosity or composition, could be incorporated as additional input features to the existing network, enabling it to learn more accurate transformations for different fluid types.

Considering gob shape, these could be integrated among the model outputs, as deadpoints would need to be adjusted to optimize shape. Two potential strategies are envisaged: first, defining “shape classes” (given the limited variability in gob shapes) and encoding them numerically; second, retaining the full shape representation as an image and transitioning to a different model architecture (e.g., a convolutional neural network). The latter approach would require a more sophisticated optimization algorithm, potentially involving adjustments along the entire Cam profile rather than only at the deadpoints, but could provide a richer and more faithful representation of the physical system.

Finally, integrating physics-based constraints or domain-specific knowledge into the learning framework could further enhance the model's reliability, particularly in situations where data alone are insufficient to guide accurate interpolation across the working space. This combination of data-driven modeling and physically-informed constraints represents a promising path toward a more robust and interpretable system.

\section{Conclusions}
\label{sec:conclusions}

In this work, we proposed a fully data-driven, AI-based control framework for optimizing the forming process in hollow glass manufacturing. Through a set of validation tests conducted on a real-world historical dataset from multiple production lines, we demonstrated the feasibility of automating the selection of machine parameters based on the desired characteristics of the final product.

The setup proved functional both computationally, running in real-time with minimal hardware requirements, and practically, delivering high accuracy in the corrections to be applied to the deadpoints. Moreover, the inversion phase enables a high level of interpretability, clearly revealing the expected plunger movements in response to product change requests.

Our analysis confirms the significant potential of deep learning-based solutions in a context where such techniques have traditionally been underutilized. More broadly, solutions of this kind can help improve production stability and efficiency, reduce waste, and enhance operator safety, marking a step toward smarter and more autonomous manufacturing processes.

\section*{Acknowledgements}
 The authors thank the Provincia Autonoma di Trento for funding this research activity, based on Legge VI, which is part of the ASTROGAME research project, led by GlassFORM.ai S.p.A..




 \bibliographystyle{elsarticle-num} 
 \bibliography{references.bib}

\begin{thebibliography}{10}
\expandafter\ifx\csname url\endcsname\relax
  \def\url#1{\texttt{#1}}\fi
\expandafter\ifx\csname urlprefix\endcsname\relax\def\urlprefix{URL }\fi
\expandafter\ifx\csname href\endcsname\relax
  \def\href#1#2{#2} \def\path#1{#1}\fi

\bibitem{https://doi.org/10.1111/ijag.15881}
Ravinder, V.~Venugopal, S.~Bishnoi, S.~Singh, M.~Zaki, H.~S. Grover, M.~Bauchy, M.~Agarwal, N.~M.~A. Krishnan, \href{https://ceramics.onlinelibrary.wiley.com/doi/abs/10.1111/ijag.15881}{Artificial intelligence and machine learning in glass science and technology: 21 challenges for the 21st century}, International Journal of Applied Glass Science 12~(3) (2021) 277--292.
\newblock \href {http://arxiv.org/abs/https://ceramics.onlinelibrary.wiley.com/doi/pdf/10.1111/ijag.15881} {\path{arXiv:https://ceramics.onlinelibrary.wiley.com/doi/pdf/10.1111/ijag.15881}}, \href {https://doi.org/https://doi.org/10.1111/ijag.15881} {\path{doi:https://doi.org/10.1111/ijag.15881}}.
\newline\urlprefix\url{https://ceramics.onlinelibrary.wiley.com/doi/abs/10.1111/ijag.15881}

\bibitem{MOURTZIS2014213}
D.~Mourtzis, M.~Doukas, D.~Bernidaki, \href{https://www.sciencedirect.com/science/article/pii/S2212827114010634}{Simulation in manufacturing: Review and challenges}, Procedia CIRP 25 (2014) 213--229, 8th International Conference on Digital Enterprise Technology - DET 2014 Disruptive Innovation in Manufacturing Engineering towards the 4th Industrial Revolution.
\newblock \href {https://doi.org/https://doi.org/10.1016/j.procir.2014.10.032} {\path{doi:https://doi.org/10.1016/j.procir.2014.10.032}}.
\newline\urlprefix\url{https://www.sciencedirect.com/science/article/pii/S2212827114010634}

\bibitem{simulation_glass_technology}
H.~Loch, D.~Krause, Mathematical Simulation in Glass Technology, 2002.
\newblock \href {https://doi.org/10.1007/978-3-642-55966-2} {\path{doi:10.1007/978-3-642-55966-2}}.

\bibitem{Yoon2025Materials}
T.~Yoon, Y.~I. Park, J.~Kim, J.-H. Kim, \href{https://doi.org/10.3390/ma18112592}{Inverse neural network approach for optimizing chemical composition in shielded metal arc weld metals}, Materials 18~(11) (2025) 2592, funding: Korea Institute of Industrial Technology (KITECH EH-25-0004) and the Ministry of Trade, Industry and Energy (NS240064).
\newblock \href {https://doi.org/10.3390/ma18112592} {\path{doi:10.3390/ma18112592}}.
\newline\urlprefix\url{https://doi.org/10.3390/ma18112592}

\bibitem{Yadav01092020}
R.~Yadav, S.~Tripathi, S.~Asati, M.~K. Das, \href{https://doi.org/10.1080/17415977.2020.1719087}{A combined neural network and simulated annealing based inverse technique to optimize the heat source control parameters in heat treatment furnaces}, Inverse Problems in Science and Engineering 28~(9) (2020) 1265--1286.
\newblock \href {http://arxiv.org/abs/https://doi.org/10.1080/17415977.2020.1719087} {\path{arXiv:https://doi.org/10.1080/17415977.2020.1719087}}, \href {https://doi.org/10.1080/17415977.2020.1719087} {\path{doi:10.1080/17415977.2020.1719087}}.
\newline\urlprefix\url{https://doi.org/10.1080/17415977.2020.1719087}

\bibitem{AstromHagg95}
K.~J. {\AA}str{\"o}m, T.~H{\"a}gglund, PID Controllers: Theory, Design, and Tuning, 2nd Edition, Instrument Society of America, Research Triangle Park, NC, USA, 1995.

\bibitem{mpc_glass_furnaces}
T.~Backx, L.~Huisman, P.~Astrid, R.~Beerkens, Model‐Based Control of Glass Melting Furnaces and Forehearths: First Principles‐Based Model of Predictive Control System Design, Vol.~24, 2008, pp. 21 -- 47.
\newblock \href {https://doi.org/10.1002/9780470294772.ch2} {\path{doi:10.1002/9780470294772.ch2}}.

\bibitem{cv_weight_gob_control}
A.~Jiménez, E.~Loinaz, F.~Rodríguez, M.~Sánchez, F.~Seco, Computer vision system for estimating and controlling the weight of glass gobs during their industrial formation process, J. Electronic Imaging 13 (2004) 613--618.
\newblock \href {https://doi.org/10.1117/1.1762888} {\path{doi:10.1117/1.1762888}}.

\bibitem{KOVACEC2010304}
M.~Kovačec, A.~Pilipović, N.~Štefanić, \href{https://www.sciencedirect.com/science/article/pii/S1755581711000216}{Improving the quality of glass containers production with plunger process control}, CIRP Journal of Manufacturing Science and Technology 3~(4) (2010) 304--310.
\newblock \href {https://doi.org/https://doi.org/10.1016/j.cirpj.2011.02.003} {\path{doi:https://doi.org/10.1016/j.cirpj.2011.02.003}}.
\newline\urlprefix\url{https://www.sciencedirect.com/science/article/pii/S1755581711000216}

\bibitem{app122111192}
A.~Bereciartua-Perez, G.~Duro, J.~Echazarra, F.~J. González, A.~Serrano, L.~Irizar, \href{https://www.mdpi.com/2076-3417/12/21/11192}{Deep learning-based method for accurate real-time seed detection in glass bottle manufacturing}, Applied Sciences 12~(21) (2022).
\newblock \href {https://doi.org/10.3390/app122111192} {\path{doi:10.3390/app122111192}}.
\newline\urlprefix\url{https://www.mdpi.com/2076-3417/12/21/11192}

\bibitem{10109368}
M.~A.~E. Latina, J.~Van Russel R. Dela~Cruz, F.~D. Delos~Santos, Empty glass bottle defect detection based on deep learning with cnn using ssd mobilenetv2 model, in: 2022 IEEE 14th International Conference on Humanoid, Nanotechnology, Information Technology, Communication and Control, Environment, and Management (HNICEM), 2022, pp. 1--6.
\newblock \href {https://doi.org/10.1109/HNICEM57413.2022.10109368} {\path{doi:10.1109/HNICEM57413.2022.10109368}}.

\bibitem{8082476}
H.~Zhang, Q.~Liu, X.~Chen, D.~Zhang, J.~Leng, A digital twin-based approach for designing and multi-objective optimization of hollow glass production line, IEEE Access 5 (2017) 26901--26911.
\newblock \href {https://doi.org/10.1109/ACCESS.2017.2766453} {\path{doi:10.1109/ACCESS.2017.2766453}}.

\bibitem{NOGRID_pointsBlow}
{NOGRID GmbH}, \href{https://www.nogrid.com/en/products/nogrid-pointsblow}{{NOGRID pointsBlow} -- simulation software for container glass forming}, accessed: 2025-08-19 (2025).
\newline\urlprefix\url{https://www.nogrid.com/en/products/nogrid-pointsblow}

\bibitem{FORGE_Glass}
{Transvalor S.A.}, \href{https://www.transvalor.com/en/industry/glass}{{FORGE\textsuperscript{\textregistered}} -- glass forming simulation software}, accessed: 2025-08-19 (2025).
\newline\urlprefix\url{https://www.transvalor.com/en/industry/glass}

\bibitem{glass_book}
E.~Le~Bourhis, \href{https://doi.org/10.1002/9783527617029}{Glass: Mechanics and Technology}, 2nd Edition, John Wiley \& Sons, Weinheim, Germany, 2014.
\newline\urlprefix\url{https://doi.org/10.1002/9783527617029}

\bibitem{glassform}
Glassform, \href{https://www.glassform.ai/}{glassform.ai}, accessed: April 11, 2025 (2025).
\newline\urlprefix\url{https://www.glassform.ai/}

\bibitem{DBLP:journals/corr/IoffeS15}
S.~Ioffe, C.~Szegedy, \href{http://arxiv.org/abs/1502.03167}{Batch normalization: Accelerating deep network training by reducing internal covariate shift}, CoRR abs/1502.03167 (2015).
\newblock \href {http://arxiv.org/abs/1502.03167} {\path{arXiv:1502.03167}}.
\newline\urlprefix\url{http://arxiv.org/abs/1502.03167}

\bibitem{DBLP:journals/corr/abs-1711-05101}
I.~Loshchilov, F.~Hutter, \href{http://arxiv.org/abs/1711.05101}{Fixing weight decay regularization in adam}, CoRR abs/1711.05101 (2017).
\newblock \href {http://arxiv.org/abs/1711.05101} {\path{arXiv:1711.05101}}.
\newline\urlprefix\url{http://arxiv.org/abs/1711.05101}

\end{thebibliography}

 \appendix

 \section{Extended evaluation on regression performance}
\label{app:extended_eval_regr}

In Section \ref{subsubsec:methods_regression}, we motivated the choice of MAE as the training metric. Still, to provide a broader and more objective assessment of the regression model, we complement the analysis reported in Table~\ref{tab:regression_results21} and Table~\ref{tab:regression_results_otherlines} with additional well-established metrics:
\begin{itemize}
    \item[-] RMSE (Root Mean Squared Error)
    \[
    RMSE = \sqrt{\frac{1}{n}\sum_{i=1}^{n}\left(y_i-\hat{y}_i\right)^2}
    \]
    \item[-] $R^2$ (coefficient of determination)
    \[
    R^2 = 1 - \frac{\sum_{i=1}^{n}\left(y_i-\hat{y}_i\right)^2}{\sum_{i=1}^{n}\left(y_i-\overline{y}_i\right)^2}
    \]
    \item[-] MedAE (median absolute error)
    \[
    MedAE = \text{Median of } |y_i-\hat{y_i}|
    \]
    \item[-] EVS (Explained Variance Score)
    \[
    EVS = 1 - \frac{Var\left(y-\hat{y}\right)}{Var(y)}
    \]
\end{itemize}
The corresponding results are summarized in Table~\ref{tab:additional_regression_metrics_weight} and Table~\ref{tab:additional_regression_metrics_length}.

\begin{table}[!ht]
\centering
\begin{tabular}{@{}r|cccc@{}}
\toprule
& \multicolumn{4}{c}{\textbf{Regression Error - Weight (validation set)}} \\
 & & & & \\
\textbf{Line ID} & \textit{RMSE [g]} & $R^2$ & \textit{MedAE [g]} & \textit{EVS} \\
\midrule
A-1 & 5.1 & 99.8 & 3.2 & 99.9 \\
A-2 & 11.9 & 98.9 & 3.8 & 98.9 \\
A-3 & 13.7 & 99.8 & 4.9 & 99.9 \\ 
B-1 & 8.9 & 98.7 & 5.2 & 98.8 \\ 
\bottomrule
\end{tabular}
\caption{Supplementary Metrics obtained from regression experiments performed on the lines A-1, A-2, A-3 and B-1, and relative to the weight estimate. Values are computed on the validation sets, which constitute the 25\% of the total available acquisitions.}
\label{tab:additional_regression_metrics_weight}
\end{table}

\begin{table}
\centering
\begin{tabular}{@{}r|cccc@{}}
\toprule
& \multicolumn{4}{c}{\textbf{Regression Error - Length (validation set)}} \\
 & & & & \\
\textbf{Line ID} & \textit{RMSE [mm]} & $R^2$ & \textit{MedAE [mm]} & \textit{EVS} \\
\midrule
A-1 & 5.1 & 98.2 & 2.4 & 98.2 \\
A-2 & 11.8 & 90.0 & 4.1 & 90.0 \\
A-3 & 9.6 & 95.6 & 4.0 & 95.5 \\ 
B-1 & 8.9 & 92.4 & 3.6 & 92.4 \\ 
\bottomrule
\end{tabular}
\caption{Supplementary Metrics obtained from regression experiments performed on the lines A-1, A-2, A-3 and B-1, and relative to the length estimate. Values are computed on the validation sets, which constitute the 25\% of the total available acquisitions.}
\label{tab:additional_regression_metrics_length}
\end{table}

\begin{figure}
    \centering
    \includegraphics[width=\linewidth]{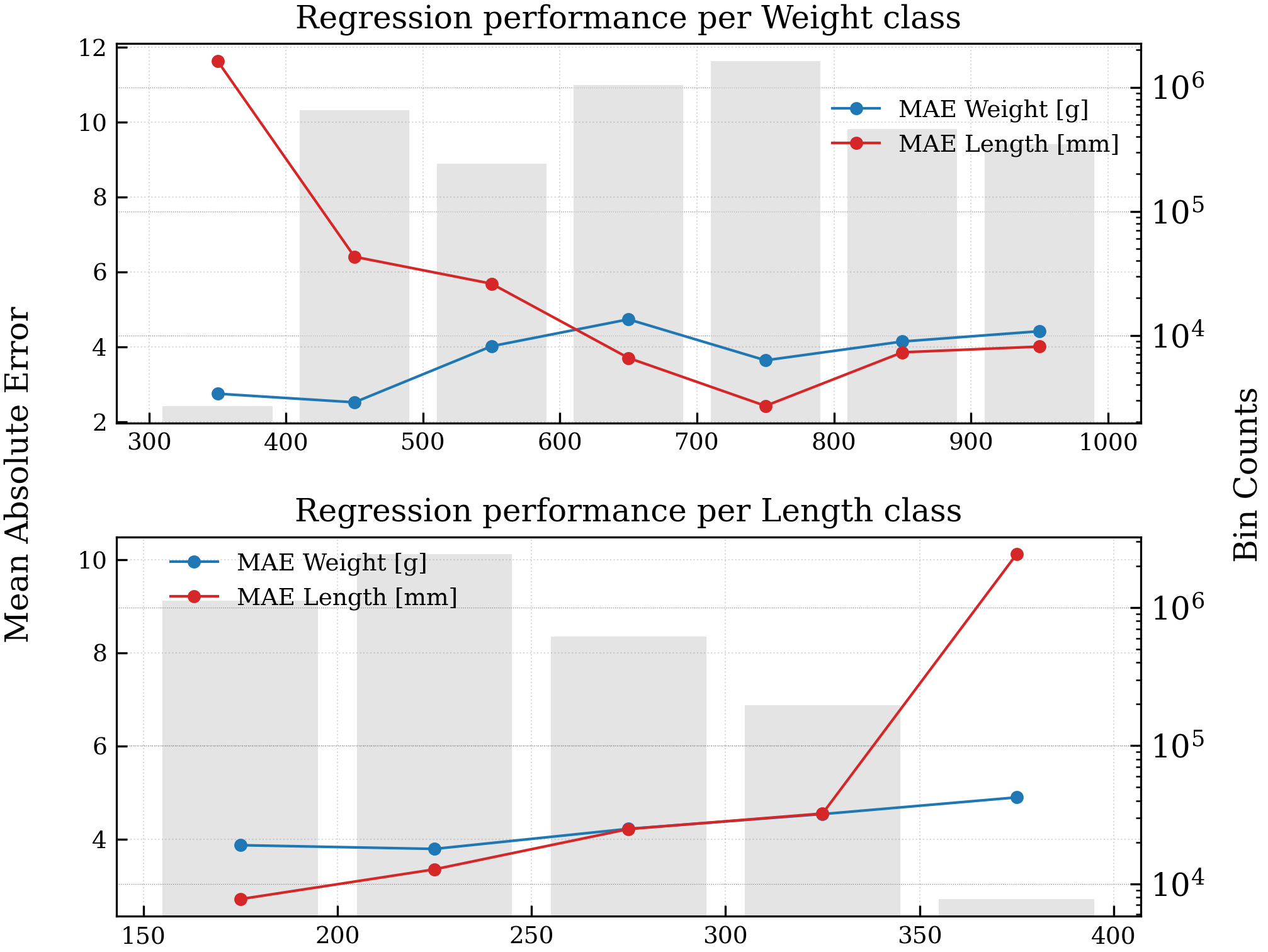}
    \caption{Comparison of reconstruction errors across weight and length classes. The background histogram shows the analysis bins and the population within each. Reported values correspond to Mean Absolute Errors computed on the validation set defined over line A-1.}
    \label{fig:regr_perf_wl_class}
\end{figure}

For the sake of completeness, an additional analysis was conducted by comparing the metrics obtained on a validation set across weight and length classes, in order to assess the presence of potential local overfitting. The results reported in Figure \ref{fig:regr_perf_wl_class} indicate that the regression performance is influenced less by the specific class itself than by the number of samples belonging to that class, which likely reflects, at least in part, the distribution present in the training dataset (as shown in \ref{fig:wl_distributions}). This finding confirms the earlier observation that the regions where the model fails to generalize correspond to those that are sparsely represented.

\section{Duplicates removal granularity}

As detailed in Sections~\ref{subsubsec:methods_regression} and~\ref{subsec:results_regression}, a major challenge in modeling was the high degree of redundancy in the dataset. Because the production process operates under stable conditions for extended periods, many input–output combinations were repeatedly observed. While natural from an industrial standpoint, such repetition provides no new information to the predictive model and may even bias the regression surface. To mitigate this, a duplicate-removal step was integrated into the preprocessing pipeline. An $n$-dimensional histogram was constructed over the joint space of input and output variables, with bin widths defined according to the nominal resolution of the measurement instruments and the accuracy of the installed sensors (see Table~\ref{tab:default_bins_dupl_rem}). Overrepresented bins were selectively pruned, reducing the training set size, often by more than 80\%. while retaining all physically meaningful variability. This histogram-based regularization offered multiple benefits: it reduced computational cost, balanced the dataset, and produced a smoother loss landscape, thereby lowering the risk of local overfitting.

\begin{table}
    \centering
    \begin{tabular}{l|c}
        \toprule
        \textbf{Variable} & \textbf{Bin size} \\
        \midrule
        Weight class [g,\%] & 0.0025\\
        Length class [mm,\%] & 0.008\\
        Temperature [$\circ$C] & 2\\
        Machine speed & 0.5\\
        Deadpoints x3 [mm] & 0.05\\
        Weight variation [g] & 0.0015\\
        Lenght variation [mm] & 0.0015\\
        Tube rotation speed & 1\\
        Shear-plunger phase & 1\\
        \bottomrule
    \end{tabular}
    \caption{Default bin sizes for the input and output variables used in the regression model. The binning strategy for Weight and Length was designed to be scalable, with bin widths defined approximately as the inverse of the measurement precision. This ensures that lower values, which require higher accuracy, are discretized more finely, whereas higher values are grouped into wider bins, reflecting the reduced need for precision at larger scales.}
    \label{tab:default_bins_dupl_rem}
\end{table}

\begin{figure}
    \centering
    \includegraphics[width=\linewidth]{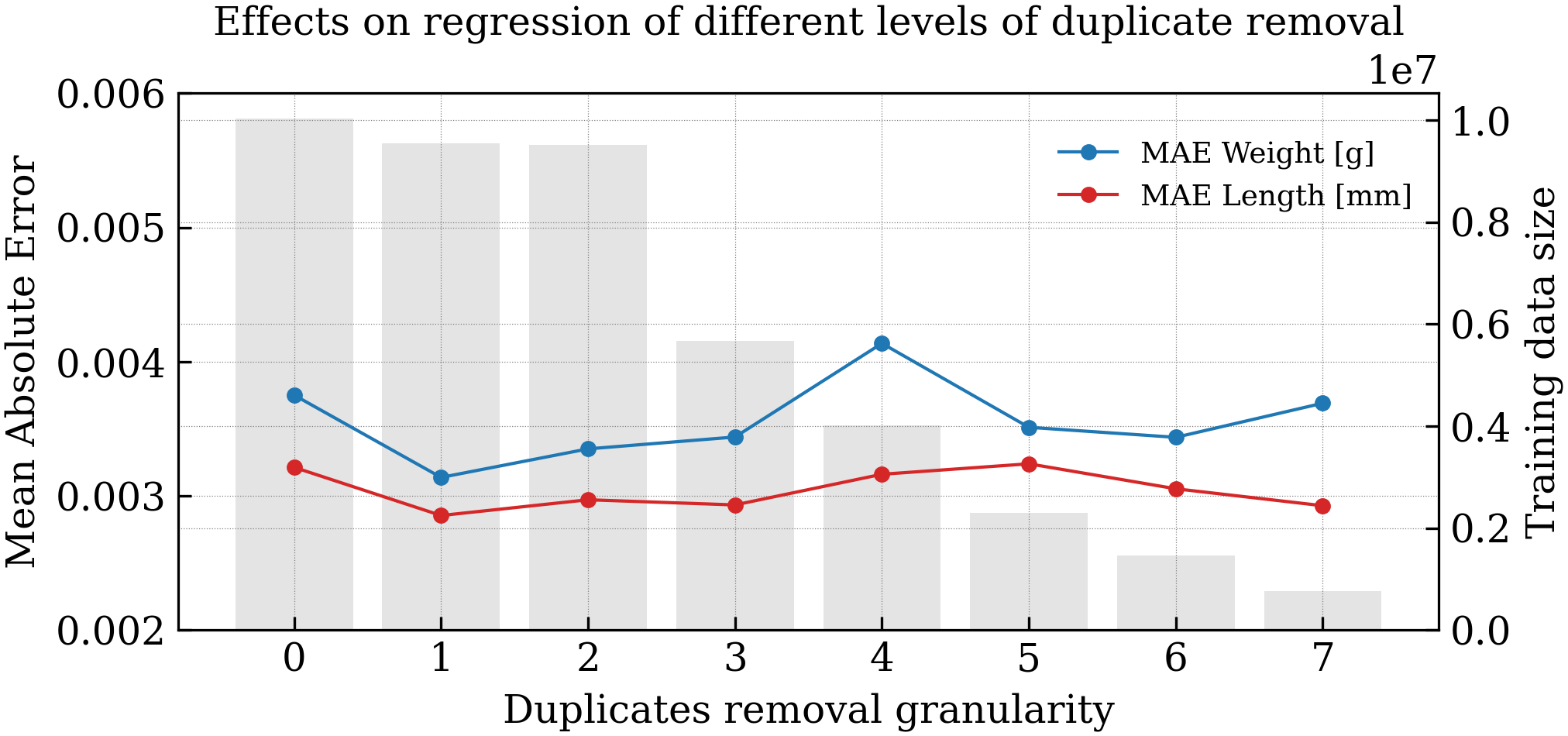}
    \caption{Effects of different levels of the duplicates removal step over the final validation metric for line A-1. The background histogram, representing the corresponding training size, is significantly reduced while the final performance metrics is only partially affected.}
    \label{fig:dupl_removal_analysis}
\end{figure}

To validate the approach, the regression model was retrained multiple times with increasingly coarse binning, corresponding to stronger subsampling. To ensure a fair comparison, duplicates were removed only from the training dataset, while the validation one remained unchanged in all runs. As shown in Figure\ref{fig:dupl_removal_analysis}, the training set size decreased progressively to about one-tenth of its original value, while the mean absolute errors (MAE) on weight and length, remained quite stable. The absence of systematic performance degradation confirms that most discarded samples were indeed redundant. Overall, the proposed histogram-based duplicate removal not only improves training speed but also enhance robustness, providing a scalable strategy for learning from highly repetitive industrial datasets.

\end{document}